\documentclass[a4paper,11pt]{article}

\usepackage{breakurl}
\usepackage[breaklinks=true, hidelinks]{hyperref}
\usepackage{breakcites}

\usepackage{amssymb,amsmath,array,bm}
\usepackage{booktabs}

\usepackage{tikz}
\usetikzlibrary{arrows}

\usepackage{pgfplots}
\pgfplotsset{compat=1.15}
\usetikzlibrary{arrows}

\usepackage[a4paper, left=3cm, right=2.5cm, top=3cm, bottom=3cm]{geometry}

\usepackage{authblk}
\usepackage{natbib}
\usepackage{doi}

\usepackage[utf8]{inputenc}
\usepackage[T1]{fontenc}
\usepackage[ngerman]{babel}
\usepackage{csquotes}

\usepackage{csquotes}
\usepackage[acronym]{glossaries}
\usepackage{graphicx}
\usepackage{subcaption}
\usepackage{wrapfig}
\newglossaryentry{formula}{name=formula,
                           description={A mathematical expression}}

\newacronym{KI}{KI}{künstliche Intelligenz}
\newacronym{ML}{ML}{maschinelles Lernen}
\newacronym{RP}{RP}{richtig-Positive}
\newacronym{FN}{FN}{falsch-Negative}
\newacronym{FP}{FP}{falsch-Positive}
\newacronym{RN}{RN}{richtig-Negative}                 
\newacronym{RPR}{RPR}{Richtig-positiv-Rate}        % TPR = TP / (TP+FN) = 1 - FNR
\newacronym{FPR}{FPR}{Falsch-positiv-Rate}         % FPR = FP / (FP+TN) = 1 - TNR
\newacronym{FNR}{FNR}{Falsch-negativ-Rate}         % FNR = FN / (TP+FN)
\newacronym{PPW}{PPW}{positiv-prädiktive-Wert}     % PPV = TP / (TP+FP) = 1 - FDR
\newacronym{NPW}{NPW}{negativ-prädiktive-Wert}     % NPV = TN / (FN+TN) = 1 - FOR
\newacronym{FTU}{FTU}{Fairness through Unawareness}
\newacronym{FTA}{FTA}{Fairness through Awareness}
\newacronym{ERC}{ERC}{European Research Council}
\newacronym{DSGVO}{DSGVO}{Datenschutz-Grundverordnung}
\newacronym{AGG}{AGG}{Allgemeine Gleichbehandlungsgesetz}
% ...
% ...
% ...
% ...
% ...
% ...
% ...
% ...
% ...
% ...
% ...
% ...

\usepackage{hyphenat}
\definecolor{FN_discriminated}{RGB}{188, 209, 247}
\definecolor{FN_privileged}{RGB}{119, 162, 242}
\definecolor{TP_discriminated}{RGB}{119, 162, 242}
\definecolor{TP_privileged}{RGB}{46, 97, 209}
\definecolor{TN_discriminated}{RGB}{252, 215, 141}
\definecolor{TN_privileged}{RGB}{247, 191, 84}
\definecolor{FP_discriminated}{RGB}{247, 191, 84}
\definecolor{FP_privileged}{RGB}{235, 159, 9}

\definecolor{grid}{rgb}{0.7529411764705882,0.7529411764705882,0.7529411764705882}

\makeindex             % used for the subject index
\title{Fairness in KI-Systemen}

\author[1]{Janine Strotherm}
\author[2]{Alissa Müller}
\author[1]{Barbara Hammer}
\author[1]{Benjamin Paaßen}
\affil[1]{Technische Fakultät, Universität Bielefeld}
\affil[2]{Medizinische Fakultät OWL, Universität Bielefeld}

\date{Vordruck ohne peer-review; Original unter \url{https://doi.org/10.1007/978-3-658-43816-6_9}}

\begin{document}

\maketitle

\pagestyle{myheadings}
\markright{Vorabdruck ohne peer-review von \cite{Strotherm2024}: Fairness in KI-Systemen}

\begin{abstract}
%Je mehr KI-gestützte Entscheidungen das Leben von Menschen betreffen, desto wichtiger ist die Fairness solcher Entscheidungen. In diesem Kapitel geben wir eine Einführung in die Forschung zu Fairness im maschinellen Lernen. Wir erklären die wesentlichen Fairness-Definitionen und Strategien zur Erreichung von Fairness anhand konkreter Beispiele und ordnen die Fairness-Forschung in den europäischen Kontext ein. Unser Beitrag richtet sich dabei an ein interdisziplinäres Publikum und verzichtet daher auf die mathematische Formulierung sondern betont Visualisierungen und Beispiele.
The more AI-assisted decisions affect people's lives, the more important the fairness of such decisions becomes. In this chapter, we provide an introduction to research on fairness in machine learning. We explain the main fairness definitions and strategies for achieving fairness using concrete examples and place fairness research in the European context. Our contribution is aimed at an interdisciplinary audience and therefore avoids mathematical formulation but emphasizes visualizations and examples.
\end{abstract}

\section{Einleitung}

Systeme der künstlichen Intelligenz (\acrshort{KI}) werden zunehmend verwendet, um Menschen bei schwierigen Entscheidungen zu unterstützen, wie etwa Gerichte bei der Beurteilung der Gefährlichkeit von Angeklagten (vgl.\ \cite{Angwin2016MachineBias}), Banken bei der Vergabe von Krediten (vgl.\ \cite{Dastile2020}) oder Unternehmen bei der Bewertung von Bewerber*innen (vgl.\ \cite{Dastin2018}). In all diesen Beispielen geht es um \emph{kritische} Entscheidungen, die den Unterschied machen, ob man eingesperrt wird, sich ein Eigenheim leisten kann, oder einen hochbezahlten Job erhält. Bei solchen \emph{kritischen} Entscheidungen ist \emph{Fairness} entscheidend (vgl.\ \cite{EuropeanComission2019TrustworthyAI,ExecutiveOffice2016BigData}). So sollte z.B.\ die Entscheidung, ob ich einen Job bekomme, von der Eignung, nicht aber vom Geschlecht abhängen.

Oberflächlich könnte man annehmen, dass ein \acrshort{KI}-System sowieso fair ist -- schließlich hat ein Automatismus keine Vorurteile und wendet das strikt gleiche Entscheidungsschema auf alle an. Tatsächlich aber können zahlreiche indirekte (und unbeabsichtigte) Effekte trotzdem dazu führen, dass \acrshort{KI}-Systeme Ergebnisse liefern, die Menschen als unfair empfinden, also beispielsweise Bewerber*innen für einen Job auf der Basis von für die Tätigkeit irrelevanten Attributen wie dem Geschlecht schlechter bewerten (vgl.\ \cite{Dastin2018,ExecutiveOffice2016BigData}). Solche automatisierte Diskriminierung ist besonders problematisch, weil sie mathematisch objektiv erscheint, viele Menschen gleichzeitig betrifft und sich aus schwer einsehbaren komplexen Berechnungen oder statistischen Modellierungen ergeben kann (vgl.\ \cite{Wachter2021}).

Ausgelöst durch einschlägige Negativbeispiele (vgl.\ \cite{Angwin2016MachineBias,Buolamwini2018GenderShades,Dastin2018}) hat sich die Fairness-Forschung zu \acrshort{KI}-Systemen zu einem eigenen Forschungsfeld mit eigenen Konferenzen entwickelt (z.B.\ der \emph{ACM Conference on Fairness, Accountability, and Transparency} FAccT). So hat sich Fairness als ein wichtiges Kriterium zur Evaluation von \acrshort{KI}-Systemen etabliert, insbesondere im europäischen Kontext (vgl.\ \cite{BringasColmenarejo2022,Calvi2023,EuropeanComission2019TrustworthyAI,Kaminski2021,Wachter2021}). Daher ist es notwendig, sich bei der Einführung von \acrshort{KI}-Systemen, die Menschen betreffen, mit Fairness auseinander zu setzen.

Dieser Text hat die Ziele, einen Überblick über die wichtigsten Konzepte zu Fairness in \acrshort{KI}-Systemen zu geben (Abschnitt~\ref{section_Fairness-Definitionen}), Strategien zu erklären, wie sich Fairness in \acrshort{KI}-Systemen erreichen lässt (Abschnitt~\ref{section_StrategienZurErreichungVonFairness}) und die Forschung in den europäischen Kontext einzubetten (Abschnitt~\ref{section_AnwendungAufDenEuropäischenKontext}). Frühere Überblicksartikel haben bereits Teile dieser Ziele adressiert (vgl.\ \cite{Barocas2019FairnessBook,Mehrabi2021SurveyOnFairness,Pessach2022ReviewOnFairness}), allerdings gibt es bisher keinen Artikel, der den Stand der \acrshort{KI}-Fairness Forschung aus einer europäischen Perspektive für ein nicht-technisches Publikum zusammenfasst. Diese Lücke schließen wir hiermit. Zunächst aber behandeln wir das nötige Hintergrundwissen für die \acrshort{KI}-Fairness-Forschung.

\section{Hintergrund}

\subsection{Grundbegriffe}
\label{subsection_Grundbegriffe}

In diesem Kapitel befassen wir uns mit künstlichen Systemen, die automatisch 
kritische Entscheidungen unterstützen und auf Daten trainiert sind. Künstliche Systeme,
die intelligentes (menschliches) Verhalten in technischen Systemen realisieren, 
werden gemeinhin als \acrshort{KI}-Systeme bezeichnet. Das Training solcher Systeme auf Daten heißt auch \gls{ML}.

Um die Grundbegriffe fassbar zu machen, führen wir sie anhand eines konkreten 
Erklärbeispiels ein, das sich durch das gesamte Kapitel ziehen wird.
Nehmen wir an, ein Konzern will Software-Entwickler*innen einstellen, erhält aber so
viele Bewerbungen, dass eine automatische Erstbegutachtung hilfreich wäre. Daher wird
ein \acrshort{KI}-System mit \gls{ML} trainiert, das Bewerbungen als Eingabe erhält
und entscheiden soll, ob es sich um eine*n geeignete*n Kandidat*in handelt oder 
nicht. Die aus den Bewerbungen extrahierten Daten wie Abschlussnoten oder Jahre an Berufserfahrung,
die das System für seine Entscheidung berücksichtigt, 
nennen wir auch \emph{Merkmale} oder 
\emph{Eingaben}. Die Entscheidungen des Systems nennen wir auch \emph{Vorhersagen}
oder \emph{Ausgaben}.
Das fertig trainierte Entscheidungsschema, das von Eingaben auf Ausgaben abbildet,
nennen wir \emph{Modell}. Weil das Modell wie in diesem Fall oft zwischen zwei diskreten
Möglichkeiten entscheiden muss (\textit{positiv} bzw. geeignet oder \textit{negativ} bzw. ungeeignet), sprechen wir dann auch von einem
binären \emph{Klassifikationsmodell} oder \emph{Klassifikator}. 
Solche Klassifikatoren funktionieren oft in zwei Schritten:
Zuerst wird ein \emph{Score} (also ein numerischer Wert) berechnet, der angeben soll, für
wie geeignet das Modell den*die Bewerber*in hält. In einem zweiten Schritt werden dann
alle Bewerber*innen als geeignet klassifiziert, deren Score jenseits eines Schwellwerts
liegt. 
Zum Beispiel könnte unser Modell einen Score auf einer Skala von 0 bis 100 Punkten
ausgeben und dann alle Bewerber*innen mit einem Wert über 80 Punkten als geeignet
klassifizieren, die übrigen als ungeeignet.

Aber was bedeutet es eigentlich, ein Modell zu \emph{trainieren}? 
Im sogenannten überwachten Lernen, einer weit verbreiteten Art des \gls{ML}, sind dazu \emph{Trainingsdaten} notwendig, für die die gewünschte Ausgabe des Modells bereits bekannt ist.
In unserem Beispiel nehmen wir an, dass der Konzern bereits Bewerbungen früherer 
Bewerber*innen gesammelt und danach sortiert hat, ob die jeweiligen Bewerber*innen abgelehnt
oder angestellt wurden. Die gewünschte Ausgabe des Modells für abgelehnte 
Bewerbungen ist \enquote{ungeeignet} und die für Bewerbungen
angestellter Bewerber*innen ist \enquote{geeignet}. Die gewünschte Antwort nennen wir auch
das \emph{Label} des \textit{Datenpunkts}\footnote{
    Die Datenpunkte können im Allgemeinen beliebigen Dingen entsprechen; wie beispielsweise einem Haus,
    wenn wir ein Modell trainieren wollen, das basierend auf Merkmalen
    des Hauses dessen Preis vorhersagen soll. In unserem Fall entspricht ein Datenpunkt aber
    immer einer Person, weshalb wir, wenn nicht zwingend anders erforderlich, auch oft direkt 
    \enquote{Person} schreiben.
}, in diesem Falle also der sich bewerbenden Person. 
Die Menge der Datenpunkte, die durch dasselbe Label ausgezeichnet sind, werden auch \emph{Klasse} genannt. 
So betrachten wir in unserem Beispiel die Klasse der geeigneten Bewerber*innen sowie die der
ungeeigneten.
Jedoch drückt das Label nicht zwingend eine unterliegende Wahrheit aus, 
sondern entspricht, wie in unserem Beispiel,
oft einem menschlichen Urteil -- dementsprechend
hat es große Auswirkungen auf die Wahrnehmung von Fairness, ob man den Labeln vertraut oder nicht.

Nachdem die Trainingsdaten vorbereitet sind, setzen wir Methoden des \gls{ML} ein, um
ein Modell so zu trainieren, dass es für möglichst viele Eingaben die gewünschte
Ausgabe (also das Label) ausgibt. Wichtig ist dabei, dass es nicht genügt, nur auf den 
Trainingsdaten die gewünschten Ausgaben zu produzieren, denn die Trainingsdaten stellen lediglich
eine (hoffentlich repräsentative) Stichprobe aus allen möglichen Datenpunkten dar.
Das Modell soll aber eigentlich für die Gesamtheit aller möglichen Datenpunkte, der sogenannten  \emph{Grundgesamtheit},
funktionieren.

\begin{figure}[h]
    \centering
    \resizebox{0.35\linewidth}{!}{\begin{tikzpicture}[line cap=round,line join=round,>=triangle 45,x=1cm,y=1cm,fill opacity=1, font=\scriptsize]
%\begin{scope}[shift={(2,0)}]
%\filldraw[fill=TN_discriminated, draw=white] (0,0) rectangle (2,2);
%\end{scope}

%grid
%\draw [color=grid,, xstep=0.5cm,ystep=0.5cm] (-2,-2) grid (2,2);

%rectangles
%upper right
\filldraw[fill=TN_discriminated, draw=white] (0,-2) rectangle (2,2);
%upper left
\filldraw[fill=FN_discriminated, draw=white] (0,-2) rectangle (-2,2); 
%lower right
%\filldraw[fill=TN_privileged, draw=white] (0,0) rectangle (2,-2);
%lower left
%\filldraw[fill=FN_privileged, draw=white] (0,0) rectangle (-2,-2);

%quarter circles (start angle : stop angle : radius)
 % upper right
\filldraw[fill=FP_discriminated, draw=white] (0,0) -- (0,-1) arc (-90:90:1) -- (0,0);
% upper left
%\filldraw[fill=TP_discriminated, draw=white] (0,0) -- (-1,0) arc (180:90:1) -- (0,0);
% lower right
\filldraw[fill=TP_discriminated, draw=white] (0,0) -- (0,-1) arc (270:90:1) -- (0,0);
% lower left
%\filldraw[fill=TP_privileged, draw=white] (0,0) -- (-1,0) arc (180:270:1) -- (0,0);

%Beschriftung
%\node at (3,1)[text width=1.6cm, align=left]{diskriminierte Gruppe};
%\node at (3,-1)[text width=1.6cm, align=left] {privilegierte Gruppe};
\node at (-1,2.25)[text width=1.6cm, align=center] {positive Klasse};
\node at (1,2.25)[text width=1.6cm, align=center] {negative Klasse};
\node at (0,0.5)[text width=1.6cm, align=center] {positive Vorhersage};
\node at (0,1.5)[text width=1.6cm, align=center] {negative Vorhersage};
\node at (-0.5,0)[text width=1.6cm, align=center] {RP};
\node at (1.5,0)[text width=1.6cm, align=center] {RN};
\node at (-1.5,0)[text width=1.6cm, align=center] {FN};
\node at (0.5,0)[text width=1.6cm, align=center] {FP};
\end{tikzpicture}}
    \caption{Darstellung möglicher Ausgänge und Fehler eines \gls{ML}-Modells.}
    \label{figure_MLGrundbegriffe}
\end{figure}
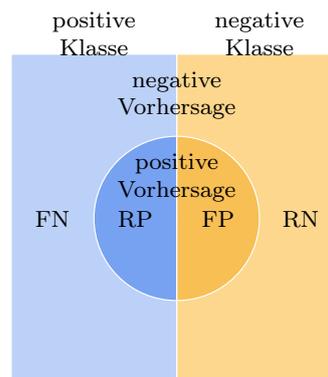

Um die Qualität des Modells auf der Grundgesamtheit abzuschätzen, nutzt man \textit{Testdaten}, 
welche dieselbe Struktur der Trainingsdaten aufweisen, jedoch nicht zum Trainieren genutzt wurden. 
Die Vorhersagen des trainierten Modells auf diesen Testdaten werden unter Einbeziehung ihrer Label 
mittels Evaluationsmaßen evaluiert. 
Solche Evaluationsmaße lassen sich am  einfachsten anhand einer Grafik beschreiben:
In Abbildung~\ref{figure_MLGrundbegriffe} werden mögliche Ausgänge der Vorhersage eines Modells
beschrieben.

\noindent
Das gesamte Bild beschreibt die betrachtete Grundgesamtheit, in unserem Beispiel also 
alle Bewerber*innen bzw.\ deren Bewerbungen.
Die linke Bildhälfte stellt die positive Klasse dar, also alle als \enquote{geeignet} gelabelte Bewerbungen,
die rechte Bildhälfte die negative Klasse, also als \enquote{ungeeignet} gelabelte Bewerbungen.
Der Kreis beschreibt alle Datenpunkte mit positiver Vorhersage, stellt also beispielsweise
alle Bewerbungen dar, die vom Modell als \enquote{geeignet} bewertet werden.
Dagegen enthält die Region außerhalb des Kreises alle Datenpunkte mit negativer Vorhersage, 
wie im Beispiel die vom Modell als \enquote{ungeeignet} bewerteten Bewerbungen. 
Somit ergeben sich vier mögliche Ausgänge des Modells:
Wir nennen die Menge an Datenpunkten, also Personen, links außerhalb des Kreises \gls{FN} 
%(sollten positiv prädiziert werden, wurden aber fälschlicherweise negativ prädiziert), 
(wurden fälschlicherweise negativ prädiziert), 
die Menge links innerhalb des Kreises \gls{RP} 
%(sollten positiv prädiziert werden und wurden auch so prädiziert), 
(wurden richtigerweise positive prädiziert), 
die Menge rechts innerhalb des Kreises \gls{FP} 
%(sollten negativ prädiziert werden, wurden aber fälschlicherweise positiv prädiziert) und 
(wurden fälschlicherweise positiv prädiziert) und 
die Menge rechts außerhalb des Kreises \gls{RN} 
%(sollten negativ prädiziert werden und wurden auch so prädiziert).
(wurden richtigerweise negativ prädiziert).

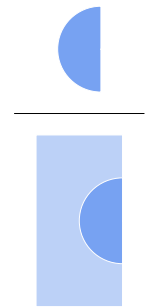
\begin{wrapfigure}{r}{0.3\textwidth}
    %\vspace{-10pt}
    \centering
    \resizebox{0.4\linewidth}{!}{
    \begin{tikzpicture}[line cap=round,line join=round,>=triangle 45,x=1cm,y=1cm,fill opacity=1]

% TODO scopes definieren um koordinaten zu verschieben
%grid
%\draw [color=grid,, xstep=0.5cm,ystep=0.5cm] (-2,-2) grid (2,2);

\begin{scope}[shift={(0.5,1.5)}]
%quarter circles (start angle : stop angle : radius)

% upper left
\filldraw[fill=TP_discriminated, draw=white] (0,0) -- (0,-1) arc (270:90:1) -- (0,0);

% lower left
%\filldraw[fill=TP_privileged, draw=white] (0,0) -- (-1,0) arc (180:270:1) -- (0,0);
\end{scope}

\draw[color=black] (-1.5,0) -- (1.5,0);

\begin{scope}[shift={(1,-2.5)}]
%rectangles

%upper left
\filldraw[fill=FN_discriminated, draw=white] (0,-2) rectangle (-2,2); 

%lower left
%\filldraw[fill=FN_privileged, draw=white] (0,0) rectangle (-2,-2);

%quarter circles (start angle : stop angle : radius)

% upper left
\filldraw[fill=TP_discriminated, draw=white] (0,0) -- (0,-1) arc (270:90:1) -- (0,0);

% lower left
%\filldraw[fill=TP_privileged, draw=white] (0,0) -- (-1,0) arc (180:270:1) -- (0,0);
\end{scope}

\end{tikzpicture}}
    \caption{Darstellung der \acrshort{RPR}.}
    \label{figure_TPR}
    \vspace{-10pt}
\end{wrapfigure}
Evaluationsmaße bilden in der Regel Verhältnisse dieses Schemas ab, 
wie beispielsweise das Verhältnis zwischen den \gls{RP} und der positiven Klasse im Falle der \gls{RPR}. 
In anderen Worten gibt die \gls{RPR} damit die Wahrscheinlichkeit an, eine positive Vorhersage zu erhalten (Datenpunkt liegt im Kreis),
gegeben, dass der Datenpunkt tatsächlich der positiven Klasse angehört 
(Datenpunkt liegt in linker Bildhälfte).
In unserem Beispiel entspricht die \gls{RPR} der Wahrscheinlichkeit, als geeignet klassifiziert zu werden,
gegeben dass die sich bewerbende Person tatsächlich geeignet ist.
Eine entsprechende Visualisierung ist in Abbildung \ref{figure_TPR} gegeben. 

Im weiteren Verlauf verwenden wir die Darstellungsformen aus Abbildung~\ref{figure_MLGrundbegriffe} und~\ref{figure_TPR}
durchgängig, um Fairness-Definitionen zu illustrieren.

\subsection{Fairness-Grundbegriffe}
\label{subsection_Fairness}

Wir betrachten Fairness in \acrshort{KI}-Systemen vor allem dahingehend, wer vom Modell
als positiv (z.B.\ \enquote{geeignet}) klassifiziert wird und wer nicht.
Dazu wird ein \emph{sensibles} bzw.\ \emph{geschütztes Merkmal} (z.B.\ Geschlecht) definiert, das die
Bevölkerung in Teilgruppen zerlegt (z.B.\ Geschlechtsidentitäten), welche gleich behandelt
werden sollen. Häufig wird eine Teilgruppe als \emph{geschützte} bzw.\ \emph{betroffene Gruppe}
definiert, die gegenüber der übrigen Bevölkerung nicht benachteiligt werden soll.
In unserem Beispiel benennen wir FLINTA* (Frauen, Lesben, intergeschlechtliche, nichtbinäre, transgeschlechtliche und 
agender-Per\-sonen, also geschlechtsbezogen und patriarchal diskriminierte Menschen)
als geschützte Gruppe und (Cis-)Männer als die übrige Bevölkerung.\footnote{
Wir weichen an dieser Stelle von der Standardverwendung der englischsprachigen 
Literatur der betrachteten Gruppen \textit{Männer} und \textit{Frauen} ab, um auch andere 
Geschlechtsidentitäten einzubeziehen.}
Unfairness könnte dann beispielsweise bedeuten, dass wesentlich weniger Bewerbungen von FLINTA* als geeignet klassifiziert werden als von Männern (vgl. Abschnitt \ref{subsection_Gruppenfairness}).

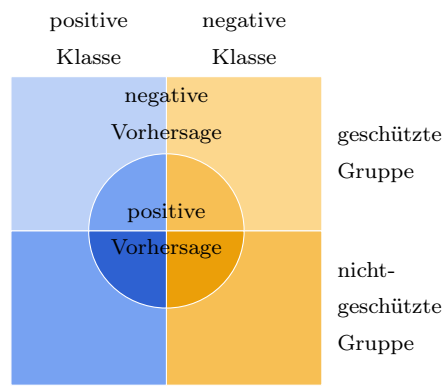
\begin{figure}[h]
    \centering
    \resizebox{0.4\linewidth}{!}{\begin{tikzpicture}[line cap=round,line join=round,>=triangle 45,x=1cm,y=1cm,fill opacity=1]
\begin{scope}[shift={(2,0)}]
%\filldraw[fill=TN_discriminated, draw=black] (0,0) rectangle (2,2);
\end{scope}

%grid
%\draw [color=grid,, xstep=0.5cm,ystep=0.5cm] (-2,-2) grid (2,2);

%rectangles
%upper right
\filldraw[fill=TN_discriminated, draw=white] (0,0) rectangle (2,2);
%upper left
\filldraw[fill=FN_discriminated, draw=white] (0,0) rectangle (-2,2); 
%lower right
\filldraw[fill=TN_privileged, draw=white] (0,0) rectangle (2,-2);
%lower left
\filldraw[fill=FN_privileged, draw=white] (0,0) rectangle (-2,-2);

%quarter circles (start angle : stop angle : radius)
 % upper right
\filldraw[fill=FP_discriminated, draw=white] (0,0) -- (1,0) arc (0:90:1) -- (0,0);
% upper left
\filldraw[fill=TP_discriminated, draw=white] (0,0) -- (-1,0) arc (180:90:1) -- (0,0);
% lower right
\filldraw[fill=FP_privileged, draw=white] (0,0) -- (1,0) arc (0:-90:1) -- (0,0);
% lower left
\filldraw[fill=TP_privileged, draw=white] (0,0) -- (-1,0) arc (180:270:1) -- (0,0);

%Beschriftung
\node at (3,1)[text width=1.6cm, align=left]{\scriptsize geschützte Gruppe};
\node at (3,-1)[text width=1.6cm, align=left] {\scriptsize nicht-geschützte Gruppe};
\node at (-1,2.5)[text width=1.6cm, align=center] {\scriptsize positive Klasse};
\node at (1,2.5)[text width=1.6cm, align=center] {\scriptsize negative Klasse};
\node at (0,0)[text width=1.6cm, align=center] {\scriptsize positive Vorhersage};
\node at (0,1.5)[text width=1.6cm, align=center] {\scriptsize negative Vorhersage};
\end{tikzpicture}}
    \caption{Darstellung möglicher Ausgänge eines \gls{ML}-Modells unter Berücksichtigung des sensiblen Merkmals.}
    \label{figure_MLGrundbegriffeInklusiveSensiblesMerkmal}
\end{figure}

Im Falle eines sensiblen Merkmals mit ebenfalls zwei Ausprägungen lässt sich dieses in das Schema nach Abbildung~\ref{figure_MLGrundbegriffe} ergänzen.
Das erweiterte Schema ist in Abbildung~\ref{figure_MLGrundbegriffeInklusiveSensiblesMerkmal} dargestellt.

\noindent
Die obere Bildhälfte repräsentiert nun die geschützte Gruppe, beispielsweise die Bewerbungen von FLINTA*, 
die untere Bildhälfte repräsentiert dagegen die übrige Bevölkerung, beispielsweise die Bewerbungen von Männern. 
Unfairness könnte nun bedeuten, dass der Halbkreis oben
(z.B.\ als geeignet klassifizierte Bewerbungen von FLINTA*) 
weniger Datenpunkte enthält als der Halbkreis unten 
(z.B.\ als geeignet klassifizierte Bewerbungen von Männern) (vgl. Abschnitt \ref{subsection_Gruppenfairness}).

Allerdings bedeutet die Abwesenheit von Unfairness zwischen Männern und FLINTA* nicht, dass das ganze System fair muss, denn auch einzelne Personen (vgl. Abschnitt \ref{subsection_IndividuelleFairness}) und andere Gruppen (Menschen verschiedener Hautfarben oder Altersklassen und Schnittmengen all dieser Gruppen) könnten immer noch unfair behandelt werden.

Außerdem muss abschließend noch erwähnt werden, dass die Abbildungen \ref{figure_MLGrundbegriffe} und \ref{figure_MLGrundbegriffeInklusiveSensiblesMerkmal} zur Visualisierung vereinfachte Darstellungen sind, die sich in der Realität gerade dadurch auszeichnen, dass abgebildete Subgruppen unterschiedlich groß sind. Dies würde in den Abbildungen unterschiedlich großen Viertelkreisen, Vierecken usw.\ entsprechen.
%bzw.\ bedeuten, dass in den entsprechenden Bereichen unterschiedlich viele Datenpunkte vorhanden sind.
Diese Ungleichheiten sind es gerade, die als Unfairness wahrgenommen werden.
Auf diese Aspekte gehen wir in Abschnitt~\ref{section_Fairness-Definitionen} ein, wo wir eine Reihe von
Fairness-Definitionen vorstellen.

\subsection{Beispiele für Unfairness}
\label{subsection_BeispieleFürUnfairness}

Die Diskussion von Fairness in \acrshort{KI}-Systemen ist durch konkrete Fallbeispiele aus der Praxis motiviert. Gründe für hierbei auftretende Unfairness können vielfältig und müssen nicht absichtlicher Natur sein. Im Alltag beschränkt sich der Diskriminierungsbegriff häufig auf \textit{direkte} Diskriminierung bzw.\ Unfairness als Folge der Verwendung sensibler Merkmale (vgl.\ \cite{Mehrabi2021SurveyOnFairness}) oder der absichtlichen Verwendung von stellvertretenden Merkmalen (vgl.\ \cite{Barocas2019FairnessBook}). Bei \acrshort{KI}-gestützten Entscheidungen dagegen ist \emph{indirekte} Diskriminierung bzw. Unfairness häufiger, d.h.\ Unfairness tritt trotz vermeintlich neutraler Regeln unabsichtlich auf (vgl.\ \cite{Pessach2022ReviewOnFairness}), beispielsweise durch pauschalisierte Gruppenstatistiken (vgl.\ \cite{Mehrabi2021SurveyOnFairness}). Letzteres beruht darauf, dass sich \gls{ML}-Methoden oft auf Korrelationen, also statistisch zu beobachtenden Zusammenhängen, aber nicht auf Kausalität stützen. Dies macht solche Methoden anfällig gegenüber statistischen Effekten in den Daten. Wenn beispielsweise jeder Fehler eines Modells gleich gewichtet wird, so fällt eine in den Trainingsdaten minder repräsentierte Gruppe automatisch weniger ins Gewicht. Wenn nun im Rahmen der Fairness-Frage wenige Datenpunkt mit positivem Label und geschützter Gruppenzugehörigkeit, also beispielsweise geeignete FLINTA*-Bewerber*innen, im Trainingsdatensatz auftreten, so erlernt das Modell möglicherweise die Korrelation einer negativen Vorhersage und der Zugehörigkeit zur geschützten Gruppe, ohne dass diese Korrelation kausal begründet werden kann.

Solche statistischen Effekte nennen wir oft \emph{Verzerrungen}, welche durch den englischen Begriff \enquote{bias} auch im Deutschen oft \emph{Bias} genannt werden. Verzerrungen können an den verschiedensten Stellen in ein \acrshort{KI}-System gelangen, beispielsweise durch wie zuvor beschrieben verzerrte Trainingsdaten, durch vermeintlich neutrale Trainingsalgorithmen, die aber beispielsweise durch ungeeignete statistische Methoden diskriminieren, oder durch die ungünstige Einbettung der Modelle in die (soziale) Umwelt. Eine ausführliche Übersicht dieser Verzerrungen bietet beispielsweise \cite{Mehrabi2021SurveyOnFairness}.

Im Folgenden wollen wir Praxisbeispiele vorstellen, in denen solche Verzerrungen und damit Unfairness zum Tragen kommen.
So orientiert sich auch unser Erklärbeispiel an der Praxis: Der US-Konzern
Amazon hatte versucht, Bewerber*innen für Stellen in der Software-Entwicklung automatisiert zu bewerten und
stellte fest, dass das System schlechtere Scores für Bewerbungen vergab, die auf Frauen hindeuteten, woraufhin die Entwicklung des Systems eingestellt wurde (vgl.\ \cite{Dastin2018}).

\cite{Buolamwini2018GenderShades} analysierten eine Reihe kommerzieller Bildklassifikatoren, die das Geschlecht einer Person anhand eines Bildes ihres Gesichts erkennen sollten. \cite{Buolamwini2018GenderShades} konnten zeigen, dass die Klassifikatoren bei Männern genauer waren als bei
Frauen und bei \textit{weißen} Personen genauer waren als bei Schwarzen Personen\footnote{
    Wir verwenden die Adjektive \textit{weiß} in kursiv und Schwarz mit großem Anfangsbuchstaben, um die Begriffe als gesellschaftliche Konstrukte zu kennzeichnen, wie im Glossar für diskriminierungssensible Sprache von Amnesty International empfohlen (vgl.\ \cite{Amnesty2023Glossar}).  
    %TODO eig. soll man original quelle von amnesty zitieren, aber die website ist nicht mehr aufrufbar
},
also jeweils geringere Fehlerraten wie die \gls{RPR} erbrachten.
Die höchsten Fehlerraten wurden auf Bildern von Schwarzen Frauen nachgewiesen.
In einer separaten Studie wiesen \cite{Srinivas2019BiasedFaceRecognition} nach, dass verschiedene kommerzielle, staatliche sowie öffentlich-zugängliche Bildklassifikatoren Kinder schlechter wiedererkennen konnten als Erwachsene.

Eines der bekanntesten Beispiele für \gls{ML}-Systeme, deren Nutzung zu unfairen Entscheidungen führen kann, ist die Software COMPAS, die im Strafrechtssystem der USA angewandt und von \cite{Angwin2016MachineBias} kritisiert wird. Das Modell berechnet einen Risiko-Score zwischen 1 (kein Risiko) und 10 (hohes Risiko) für Angeklagte in den USA.
Dieser Risiko-Score kann zur Beurteilung solcher Angeklagten, beispielsweise als Entscheidungshilfe innerhalb eines Prozesses, herangezogen werden. Die Merkmale zur Berechnung des Scores sind Antworten auf einen Fragebogen, die teils aus den Akten der Person gezogen werden (z.B.\ Vorstrafenregister) und teils von den Angeklagten beantwortet werden. Obwohl die Hautfarbe selbst nicht abgefragt wird, konnten \cite{Angwin2016MachineBias} eine Diskriminierung gegen Schwarze als betroffene Gruppe feststellen. Der Klassifikator verursacht zwar in etwa gleich viele Fehler innerhalb der betrachteten Gruppen, allerdings unterscheidet sich die Art der Fehler zwischen den Gruppen: 
Schwarze Angeklagte werden fast doppelt so häufig fälschlich mit einem hohen Risiko bewertet (\gls{FP}) wie \textit{weiße} Angeklagte; und \textit{weiße} Angeklagte werden häufiger fälschlich mit niedrigem Risiko bewertet (\gls{FN}) als Schwarze Angeklagte.
Da hier die Konsequenz einer positiven Klassifizierung, in diesem Fall also eines hohen Risiko-Scores, strafend ist (höhere Wahrscheinlichkeit für schwerere Urteile), sind die Konsequenzen der höheren Anzahl an falsch-Positiven (\acrshort{FP}) für Schwarze deutlich schlimmer als die Konsequenzen der höheren Anzahl an falsch-Negativen (\acrshort{FN}) für \textit{weiße} Angeklagte. Dementsprechend unterstreicht dieses Beispiel, wie wichtig es ist, sich über die Konsequenzen der verschiedenen Fehlerarten bewusst zu werden und zwischen ihnen zu unterscheiden.

Diese Beispiele zeigen, wie essentiell wichtig die Betrachtung von Fairness in \gls{ML}-Systemen ist. Im folgenden Abschnitt zeigen wir, wie die Fairness solcher Systeme gemessen werden kann.

\section{Fairness-Definitionen}
\label{section_Fairness-Definitionen}

Wie die vorangegangenen Negativbeispiele (vgl.\ Abschnitt~\ref{subsection_BeispieleFürUnfairness}) zeigen, stellen auf vermeintlich neutralen \acrshort{KI}-Systemen gestützte Entscheidungen ein Problem dar, vor allem dann, wenn diese Entscheidungen bedeutende Auswirkungen auf die betroffenen Individuen haben. Viele dieser vorgestellten Studien beruhen auf zunächst scheinbar willkürlich ausgewählten Analysen mit dem Resultat, dass das analysierte Modell für die geschützte Gruppe schlechtere Ergebnisse erzielt als für die übrige Bevölkerung (vgl.\ \cite{Angwin2016MachineBias,Buolamwini2018GenderShades,Dastin2018,Srinivas2019BiasedFaceRecognition}). Tatsächlich gehören Methoden wie das Vergleichen der \glspl{RPR} zwischen Gruppen aber zu einer breiten Auswahl an Fairness-Definitionen, welche im Folgenden erläutert werden.
Solche lassen sich basierend auf verschiedenen Motivationen zunächst in vier Kategorien aufteilen: \emph{Gruppenfairness} zielt darauf ab, Gruppen gleich zu behandeln (vgl.\ \cite{Mehrabi2021SurveyOnFairness}), während \emph{individuelle Fairness} sich darauf konzentriert, ähnliche Individuen ähnlich zu behandeln (vgl.\ \cite{Dwork2012}). 
Dagegen misst \emph{kausale Fairness}, wie sich eine Änderung des sensiblen Merkmals auf eine Eingabe \emph{kausal} auswirkt und damit beispielsweise die Frage beantwortet, ob man als Software-Entwickler*in eingestellt werden würde, wenn man nicht FLINTA*, sondern ein Mann wäre. \textit{Dynamische Fairness} schließlich befasst sich vor allem damit, ob ein Klassifikator auch dann noch fair ist, wenn man (unbeabsichtigte) Rückkopplungseffekte über die Zeit mit einbezieht. Typische Fragestellungen wären, ob es vielleicht langfristig dazu führt, dass sich immer weniger FLINTA* auf Software-Entwicklungsstellen bewerben, wenn man den Status Quo fortschreibt.

Zu jeder dieser vier Kategorien gehören eine Menge an unterschiedlich motivierten Fairness-Definitionen, von welchen in den folgenden Abschnitten \ref{subsection_Gruppenfairness} bis \ref{subsection_DynamischeFairness} die wichtigsten jeder Art vorgestellt werden. Abschließend wird in Abschnitt \ref{subsection_AuswahlVonFairness-Definitionen} diskutiert, wieso unter diesen eine Auswahl getroffen werden muss und wie die richtige Auswahl für den Anwendungsfall entsprechend aus der Vielfalt an verschiedenen Fairness-Definitionen getroffen werden kann.

\subsection{Gruppenfairness}
\label{subsection_Gruppenfairness}

Die Grundidee der Gruppenfairness-Definitionen ist der Vergleich von Erfolgs- oder Fehlerraten zwischen der geschützten Gruppe und der übrigen Bevölkerung (vgl.\ \cite{Castelnovo2022ClarificationOnFairness,Mehrabi2021SurveyOnFairness,Pessach2022ReviewOnFairness,Ruf2021RightKindOfFairness}).
Ein Beispiel einer Fehlerrate, nämlich die \gls{RPR}, haben wir bereits in Abschnitt \ref{subsection_Grundbegriffe} erläutert. 
Weitere Erfolgs- sowie Fehlerraten werden an entsprechender Stelle eingeführt.
Je nachdem, welche Erfolgs- oder Fehlerrate in Betracht gezogen wird, unterscheidet man in der Gruppenfairness zwischen den Unterkategorien \textit{Unabhängigkeit}, \textit{Separation} und \textit{Suffizienz} (vgl.\ \cite{Barocas2019FairnessBook,Castelnovo2022ClarificationOnFairness,Ruf2021RightKindOfFairness}).
\textit{Unabhängigkeit} verlangt, dass die Modellausgabe unabhängig von der Gruppenzugehörigkeit ist, also dass beispielsweise die Klassifikation, ob eine Person für eine Software-Entwicklungsstelle geeignet ist, nicht davon abhängt, ob letztere männlich oder FLINTA* ist. 
Der Unabhängigkeits-Begriff ist hierbei mathematischer Natur, der sich auf die Gleichheit von Wahrscheinlichkeiten bezieht. 
Beispielsweise soll die Wahrscheinlichkeit, als geeignet klassifiziert zu werden, für FLINTA* und für Männer gleich groß sein (vgl.\ Abschnitt \ref{subsubsection_Unabhängigkeit}).
\textit{Separation} fordert ebenfalls solch eine mathematische Unabhängigkeit, allerdings nur für Daten des gleichen Label. 
Ein Beispiel dafür ist, dass die Wahrscheinlichkeit für positiv gelabelte Bewerbungen, auch vom Modell als geeignet klassifiziert zu werden, für FLINTA* und Männer gleich groß sein soll (vgl.\ Abschnitt \ref{subsubsection_Separation}).
\textit{Suffizienz} fordert dagegen eine andere Art der mathematischen Unabhängigkeit, nämlich die zwischen Label und der Gruppenzugehörigkeit, allerdings nur für Daten der gleichen Modellausgabe. 
Ein Beispiel dafür ist, dass die Wahrscheinlichkeit für als geeignet klassifizierte Bewerbungen, auch positiv gelabelt zu sein, für FLINTA* und Männer gleich groß sein soll (vgl.\ Abschnitt \ref{subsubsection_Suffizienz}).

In der Praxis sind die Wahrscheinlichkeiten, auf die sich die Definitionen beziehen, nicht bekannt, sondern müssen empirisch aus Daten geschätzt werden. Das wird durch die Messung von Erfolgs- (Unabhängigkeit) und Fehlerraten (Separation und Suffizienz) erreicht.
Im Folgenden gehen wir auf die drei Unterkategorien der Reihe nach ein und erläutern die jeweils wichtigsten Fairness-Definitionen.

\subsubsection{Unabhängigkeit}
\label{subsubsection_Unabhängigkeit}

Eine bekannte auf dem Konzept der Unabhängigkeit beruhende Fairness-Definition ist \textit{Disparate Impact}. 
Diese ist an die gesetzliche Definition des Disparate Impact aus dem US amerikanischen Raum angelehnt (vgl.\ \cite{Pessach2022ReviewOnFairness}), welche besagt, dass nach der \enquote{80$\%$ Regel} die \textit{passing rate} der einen Gruppen minimal 80$\%$ der anderen Gruppe entsprechen soll (vgl.\ \cite{Biddle2006LegalDefinition_DisparateImpact}). 
Die passing rate, zu deutsch \textit{Erfolgsrate}, gibt den relativen Anteil an positiven Vorhersagen, genauer also die an der Grundgesamtheit gemessene Wahrscheinlichkeit, zu welcher das betrachtete Modell eine positive Entscheidung trifft, an. 
Die gleichnamige Fairness-Definition fordert, dass diese Erfolgsraten über die betrachteten Gruppen hinweg gleich sind (vgl.\ \cite{Pessach2022ReviewOnFairness}).

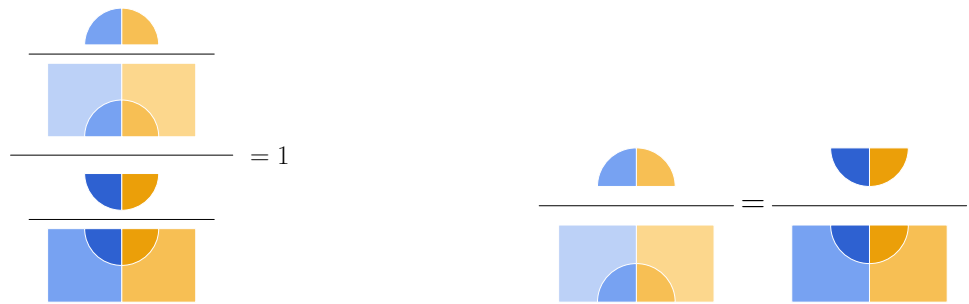
\begin{figure}[htb]
    \begin{subfigure}[t]{.49\textwidth}
        \centering
        \resizebox{0.5\linewidth}
        {!}{\begin{tikzpicture}[line cap=round,line join=round,>=triangle 45,x=1cm,y=1cm,fill opacity=1]

%grid
%\draw [color=grid,, xstep=0.5cm,ystep=0.5cm] (-3,-4.5) grid (3,4.5);

\begin{scope}[shift={(0,3)}]
%quarter circles (start angle : stop angle : radius)
 % upper right
\filldraw[fill=FP_discriminated, draw=white] (0,0) -- (1,0) arc (0:90:1) -- (0,0);
% upper left
\filldraw[fill=TP_discriminated, draw=white] (0,0) -- (-1,0) arc (180:90:1) -- (0,0);
\end{scope}

\draw[color=black] (-2.5,2.75) -- (2.5,2.75);

\begin{scope}[shift={(0,0.5)}]
%rectangles
%upper right
\filldraw[fill=TN_discriminated, draw=white] (0,0) rectangle (2,2);
%upper left
\filldraw[fill=FN_discriminated, draw=white] (0,0) rectangle (-2,2); 

%quarter circles (start angle : stop angle : radius)
 % upper right
\filldraw[fill=FP_discriminated, draw=white] (0,0) -- (1,0) arc (0:90:1) -- (0,0);
% upper left
\filldraw[fill=TP_discriminated, draw=white] (0,0) -- (-1,0) arc (180:90:1) -- (0,0);
\end{scope}

\draw[color=black] (-3,0) -- (3,0);

\begin{scope}[shift={(0,-0.5)}]
%quarter circles (start angle : stop angle : radius)
% lower right
\filldraw[fill=FP_privileged, draw=white] (0,0) -- (1,0) arc (0:-90:1) -- (0,0);
% lower left
\filldraw[fill=TP_privileged, draw=white] (0,0) -- (-1,0) arc (180:270:1) -- (0,0);
\end{scope}

\draw[color=black] (-2.5,-1.75) -- (2.5,-1.75);

\begin{scope}[shift={(0,-2)}]
%rectangles
%lower right
\filldraw[fill=TN_privileged, draw=white] (0,0) rectangle (2,-2);
%lower left
\filldraw[fill=FN_privileged, draw=white] (0,0) rectangle (-2,-2);

%quarter circles (start angle : stop angle : radius)
% lower right
\filldraw[fill=FP_privileged, draw=white] (0,0) -- (1,0) arc (0:-90:1) -- (0,0);
% lower left
\filldraw[fill=TP_privileged, draw=white] (0,0) -- (-1,0) arc (180:270:1) -- (0,0);
\end{scope}
\node at (4,0) {\huge $= 1$};  
  
\end{tikzpicture}}
        \caption{Visualisierung der Gruppenfairness-Definition Disparate Impact.}
        \label{figure_Gruppenfairness_DisparateImpact}
    \end{subfigure}
    \hfill
    \begin{subfigure}[t]{.49\textwidth}
        \centering
        \resizebox{0.75\linewidth}
        {!}{\begin{tikzpicture}[line cap=round,line join=round,>=triangle 45,x=1cm,y=1cm,fill opacity=1]

%grid
%\draw [color=grid,, xstep=0.5cm,ystep=0.5cm] (-3,-6) grid (3,6);
\begin{scope}[shift={(-3,0)}]
\begin{scope}[shift={(0,0.5)}]
%quarter circles (start angle : stop angle : radius)
 % upper right
\filldraw[fill=FP_discriminated, draw=white] (0,0) -- (1,0) arc (0:90:1) -- (0,0);
% upper left
\filldraw[fill=TP_discriminated, draw=white] (0,0) -- (-1,0) arc (180:90:1) -- (0,0);
\end{scope}

\draw[color=black] (-2.5,0) -- (2.5,0);

\begin{scope}[shift={(0,-2.5)}]
%rectangles
%upper right
\filldraw[fill=TN_discriminated, draw=white] (0,0) rectangle (2,2);
%upper left
\filldraw[fill=FN_discriminated, draw=white] (0,0) rectangle (-2,2); 

%quarter circles (start angle : stop angle : radius)
 % upper right
\filldraw[fill=FP_discriminated, draw=white] (0,0) -- (1,0) arc (0:90:1) -- (0,0);
% upper left
\filldraw[fill=TP_discriminated, draw=white] (0,0) -- (-1,0) arc (180:90:1) -- (0,0);
\end{scope}
\end{scope}

\node at (0,0) {\Huge $=$};

\begin{scope}[shift={(3,0)}]
\begin{scope}[shift={(0,1.5)}]
%quarter circles (start angle : stop angle : radius)
% lower right
\filldraw[fill=FP_privileged, draw=white] (0,0) -- (1,0) arc (0:-90:1) -- (0,0);
% lower left
\filldraw[fill=TP_privileged, draw=white] (0,0) -- (-1,0) arc (180:270:1) -- (0,0);
\end{scope}

\draw[color=black] (-2.5,0) -- (2.5,0);

\begin{scope}[shift={(0,-0.5)}]
%rectangles
%lower right
\filldraw[fill=TN_privileged, draw=white] (0,0) rectangle (2,-2);
%lower left
\filldraw[fill=FN_privileged, draw=white] (0,0) rectangle (-2,-2);

%quarter circles (start angle : stop angle : radius)
% lower right
\filldraw[fill=FP_privileged, draw=white] (0,0) -- (1,0) arc (0:-90:1) -- (0,0);
% lower left
\filldraw[fill=TP_privileged, draw=white] (0,0) -- (-1,0) arc (180:270:1) -- (0,0);
\end{scope}
\end{scope} 
  
\end{tikzpicture}}
        \caption{Visualisierung der Gruppenfairness-Definition Demographic Parity.}
        \label{figure_Gruppenfairness_DemographicParity}
    \end{subfigure}
    \caption{Visualisierung der äquivalenten Gruppenfairness-Definitionen Disparate Impact und Demographic Parity.}
\end{figure}

Eine äquivalente Fairness-Definition ist \textit{Demographic Parity}, auch \textit{Statistical Parity} genannt. 
Während auch hier die Gleichheit der Erfolgsraten über die betrachteten Gruppen hinweg verlangt wird, unterscheidet sich lediglich die mathematische Berechnung des Maßes:
Während Demographic Parity einfach die Gleichheit der beiden Erfolgsraten fordert (vgl.\ Abb. \ref{figure_Gruppenfairness_DemographicParity}), gibt Disparate Impact die Reihenfolge der Gruppen im Bruch vor (vgl.\ Abb. \ref{figure_Gruppenfairness_DisparateImpact}). Das Ergebnis des Bruchs ist größer, je höher die Erfolgsrate der geschützten Gruppe ist und sollte -- entsprechend des US amerikanischen Grundsatzes -- mindestens 0.8 betragen (vgl.\ \cite{Pessach2022ReviewOnFairness}). Beträgt der Bruch gerade eins, sind beide Formulierungen äquivalent.
Entsprechende Visualisierungen der beiden äquivalenten Fairness-Definitionen sind in Abbildung \ref{figure_Gruppenfairness_DisparateImpact} sowie \ref{figure_Gruppenfairness_DemographicParity} zu sehen.

Nach beiden Definitionen würde das betroffene Modell in unserem laufenden Beispiel also jeweils als fair bezeichnet werden, wenn der gleiche relative Anteil an FLINTA* sowie Männern als geeignet klassifiziert würde, beispielsweise jeweils 50$\%$ aller FLINTA* sowie 50$\%$ aller männlichen Bewerber*innen.

\begin{wrapfigure}{r}{0.4\textwidth}
    \vspace{-10pt}
    \centering
    \resizebox{0.9\linewidth}{!}{
    \begin{tikzpicture}[line cap=round,line join=round,>=triangle 45,x=1cm,y=1cm,fill opacity=1]

%grid
%\draw [color=grid,, xstep=0.5cm,ystep=0.5cm] (-3,-6) grid (3,6);
\begin{scope}[shift={(-1.5,0)}]
\begin{scope}[shift={(0,-0.5)}]
%quarter circles (start angle : stop angle : radius)
 % upper right
\filldraw[fill=FP_discriminated, draw=white] (0,0) -- (1,0) arc (0:90:1) -- (0,0);
% upper left
\filldraw[fill=TP_discriminated, draw=white] (0,0) -- (-1,0) arc (180:90:1) -- (0,0);
\end{scope}
\end{scope}

\node at (0,0) {\large $=$};

\begin{scope}[shift={(1.5,0)}]
\begin{scope}[shift={(0,0.5)}]
%quarter circles (start angle : stop angle : radius)
% lower right
\filldraw[fill=FP_privileged, draw=white] (0,0) -- (1,0) arc (0:-90:1) -- (0,0);
% lower left
\filldraw[fill=TP_privileged, draw=white] (0,0) -- (-1,0) arc (180:270:1) -- (0,0);
\end{scope}
\end{scope} 
  
\end{tikzpicture}}
    \caption{Visualisierung der Gruppenfairness-Definition Equal Selection Parity.}
    \label{figure_Gruppenfairness_EqualSelectionParity}
    \vspace{-10pt}
\end{wrapfigure}
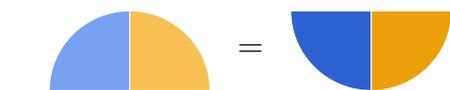
\noindent
Einen echten Unterschied stellt die Fairness-Definition namens \textit{Equal Selection Parity} dar: Hier sollen nicht Erfolgsraten, also die \text{relativen} Anteile an positiven Vorhersagen, über die betrachteten Gruppen hinweg gleich sein, sondern wie in Abbildung \ref{figure_Gruppenfairness_EqualSelectionParity} dargestellt die \textit{absoluten} Anteile an positiven Vorhersagen (vgl.\ \cite{Ruf2021RightKindOfFairness}). 

In unserem Erklärbeispiel würde das Modell entsprechend als fair bezeichnet werden, wenn gleich viele FLINTA* wie Männer als geeignet klassifiziert würden, beispielsweise jeweils 20 FLINTA* und 20 Männer.
\\
\\
Die bisherigen Fairness-Definitionen sind vor allem dadurch charakterisiert, dass sie die Label \textit{nicht} berücksichtigen. Das nicht-Nutzen der Label verleiht den Fairness-Definitionen der Kategorie Unabhängigkeit einen normativen Charakter, auf welchen wir in Abschnitt \ref{subsection_AuswahlVonFairness-Definitionen} tiefer eingehen.
Wesentlicher Nachteil dieser Definitionen ist allerdings, dass selbst ein Klassifikator, der immer die (gemäß Label) gewünschte Ausgabe erzeugt, als unfair bewertet werden kann (vgl.\ \cite{Pessach2022ReviewOnFairness}).

\subsubsection{Separation}
\label{subsubsection_Separation}

Im Gegensatz zu den Fairness-Definitionen der Kategorie Unabhängigkeit beziehen die Definitionen der Kategorien Separation und Suffizienz die Label mit ein.
Im Falle der Separation ist die zugrundeliegende Grundgesamtheit, auf welcher dann statistische Maße berechnet werden, die Menge aller Datenpunkte einer bestimmten Klasse, also beispielsweise alle Bewerber*innen, die laut Label tatsächlich geeignet sind (anstatt alle Bewerber*innen).
Typische, von der Fairness-Fragestellung zunächst losgelöste Evaluationsmaße bzw.\ \textit{Fehlerraten} eines binären Klassifikators sind hier beispielsweise die in Abschnitt \ref{subsection_Grundbegriffe} eingeführte \gls{RPR} sowie die \gls{FPR}. 
Zur Erinnerung: 
Die \gls{RPR} entspricht der Wahrscheinlichkeit, dass ein Datenpunkt, der laut Label der positiven Klasse angehört, auch korrekt in diese klassifiziert wird (vgl.\ Abb.\ \ref{figure_TPR}). 
Die zugrundeliegende Grundgesamtheit entspricht hier der linken Seite, also der positiven Klasse, aus unserem Schema aus Abbildung \ref{figure_MLGrundbegriffe}.

\begin{wrapfigure}{r}{0.3\textwidth}
    \vspace{-10pt}
    \centering
    \resizebox{0.5\linewidth}{!}{
    \begin{tikzpicture}[line cap=round,line join=round,>=triangle 45,x=1cm,y=1cm,fill opacity=1]

%grid
%\draw [color=grid,, xstep=0.5cm,ystep=0.5cm] (-2,-2) grid (2,2);

\begin{scope}[shift={(-0.5,1.5)}]

%quarter circles (start angle : stop angle : radius)
 % upper right
\filldraw[fill=FP_discriminated, draw=white] (0,0) -- (0,1) arc (90:0:1) -- (0,0);

% lower right
\filldraw[fill=FP_privileged, draw=white] (0,0) -- (1,0) arc (0:-90:1) -- (0,0);

\end{scope}

\draw[color=black] (-2,0) -- (2,0);

\begin{scope}[shift={(-1,-2.5)}]
%rectangles
%upper right
\filldraw[fill=TN_discriminated, draw=white] (0,0) rectangle (2,2);

%lower right
\filldraw[fill=TN_privileged, draw=white] (0,0) rectangle (2,-2);

%quarter circles (start angle : stop angle : radius)
 % upper right
\filldraw[fill=FP_discriminated, draw=white] (0,0) -- (0,1) arc (90:0:1) -- (0,0);
% lower right
\filldraw[fill=FP_privileged, draw=white] (0,0) -- (1,0) arc (0:-90:1) -- (0,0);

\end{scope}

\end{tikzpicture}}
    \caption{Darstellung der \acrshort{FPR}, bereits inklusive Visualisierung der betrachteten Gruppen.}
    \label{figure_FPR}
    \vspace{-10pt}
\end{wrapfigure}
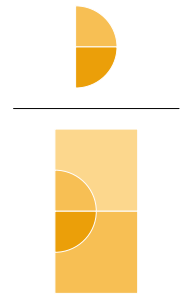
Die \gls{FPR} entspricht dagegen der Wahrscheinlichkeit, dass ein Datenpunkt, der laut Label der negativen Klasse angehört, inkorrekt in die positive Klasse klassifiziert wird (vgl.\ Abb.\ \ref{figure_FPR}).
In diesem Fall entspricht die zugrundeliegende Grundgesamtheit der rechten Seite, also der negativen Klasse, aus unserem Schema aus Abbildung \ref{figure_MLGrundbegriffe}.
Unterteilt man diese beiden zugrundeliegenden Grundgesamtheiten nun zusätzlich noch entsprechend der betrachteten Gruppen gemäß dem Schema in Abbildung \ref{figure_MLGrundbegriffeInklusiveSensiblesMerkmal} (oben bzw.\ unten), so kann man die \gls{RPR} sowie die \gls{FPR} innerhalb dieser Gruppen betrachten, woraus verschiedene Fairness-Definitionen entstehen:

\textit{Equalized Odds} ist durch den Nachteil des Unabhängigkeit-Konzepts motiviert, dass auch ein perfekter Klassifikator unfair sein kann. Unter Einbeziehung der Label kann diese und folgende Fairness-Definitionen so dem zentralem Ziel des \gls{ML}, nämlich akkurate Klassifikatoren zu entwickeln, gerecht werden (vgl.\ \cite{Hardt2016FairClassification_PostProcess_RetrainModel}).
Equalized Odds fordert die Gleichheit der \glspl{RPR} sowie der \glspl{FPR} über die betrachteten Gruppen hinweg (vgl.\ \cite{Mehrabi2021SurveyOnFairness}).

\begin{figure}[htb]
    \begin{subfigure}[t]{.49\textwidth}
        \centering
        \resizebox{0.7\linewidth}
        {!}{\begin{tikzpicture}[line cap=round,line join=round,>=triangle 45,x=1cm,y=1cm,fill opacity=1]

%grid
%\draw [color=grid,, xstep=0.5cm,ystep=0.5cm] (-3,-6) grid (3,6);
\begin{scope}[shift={(-2,0)}]
\begin{scope}[shift={(0.5,0.5)}]
%quarter circles (start angle : stop angle : radius)

% upper left
\filldraw[fill=TP_discriminated, draw=white] (0,0) -- (-1,0) arc (180:90:1) -- (0,0);
\end{scope}

\draw[color=black] (-1.5,0) -- (1.5,0);

\begin{scope}[shift={(1,-2.5)}]
%rectangles

%upper left
\filldraw[fill=FN_discriminated, draw=white] (0,0) rectangle (-2,2); 

%quarter circles (start angle : stop angle : radius)

% upper left
\filldraw[fill=TP_discriminated, draw=white] (0,0) -- (-1,0) arc (180:90:1) -- (0,0);
\end{scope}
\end{scope}

\node at (0,0) {\huge $=$};

\begin{scope}[shift={(2,0)}]
\begin{scope}[shift={(0.5,1.5)}]
%quarter circles (start angle : stop angle : radius)

% lower left
\filldraw[fill=TP_privileged, draw=white] (0,0) -- (-1,0) arc (180:270:1) -- (0,0);
\end{scope}

\draw[color=black] (-1.5,0) -- (1.5,0);

\begin{scope}[shift={(1,-0.5)}]
%rectangles

%lower left
\filldraw[fill=FN_privileged, draw=white] (0,0) rectangle (-2,-2);

%quarter circles (start angle : stop angle : radius)

% lower left
\filldraw[fill=TP_privileged, draw=white] (0,0) -- (-1,0) arc (180:270:1) -- (0,0);
\end{scope}
\end{scope} 
  
\end{tikzpicture}}
        \caption{Visualisierung der Gruppenfairness-Definition Equal Opportunity.}
        \label{figure_Gruppenfairness_EqualOpportunity}
    \end{subfigure}
    \hfill
    \begin{subfigure}[t]{.49\textwidth}
        \centering
        \resizebox{0.7\linewidth}
        {!}{\begin{tikzpicture}[line cap=round,line join=round,>=triangle 45,x=1cm,y=1cm,fill opacity=1]

%grid
%\draw [color=grid,, xstep=0.5cm,ystep=0.5cm] (-3,-6) grid (3,6);
\begin{scope}[shift={(-2,0)}]
\begin{scope}[shift={(-0.5,0.5)}]
%quarter circles (start angle : stop angle : radius)
 % upper right
\filldraw[fill=FP_discriminated, draw=white] (0,0) -- (1,0) arc (0:90:1) -- (0,0);

\end{scope}

\draw[color=black] (-1.5,0) -- (1.5,0);

\begin{scope}[shift={(-1,-2.5)}]
%rectangles
%upper right
\filldraw[fill=TN_discriminated, draw=white] (0,0) rectangle (2,2);

%quarter circles (start angle : stop angle : radius)
 % upper right
\filldraw[fill=FP_discriminated, draw=white] (0,0) -- (1,0) arc (0:90:1) -- (0,0);

\end{scope}
\end{scope}

\node at (0,0) {\huge $=$};

\begin{scope}[shift={(2,0)}]
\begin{scope}[shift={(-0.5,1.5)}]
%quarter circles (start angle : stop angle : radius)
% lower right
\filldraw[fill=FP_privileged, draw=white] (0,0) -- (1,0) arc (0:-90:1) -- (0,0);

\end{scope}

\draw[color=black] (-1.5,0) -- (1.5,0);

\begin{scope}[shift={(-1,-0.5)}]
%rectangles
%lower right
\filldraw[fill=TN_privileged, draw=white] (0,0) rectangle (2,-2);

%quarter circles (start angle : stop angle : radius)
% lower right
\filldraw[fill=FP_privileged, draw=white] (0,0) -- (1,0) arc (0:-90:1) -- (0,0);

\end{scope}
\end{scope} 
  
\end{tikzpicture}}
        \caption{Visualisierung der Gruppenfairness-Definition Predictive Equality.}
         \label{figure_Gruppenfairness_PredictiveEquality}
    \end{subfigure}
    \caption{Visualisierung der Gruppenfairness-Definition Equalized Odds.}
    \label{figure_Gruppenfairness_EqualizedOdds}
\end{figure}
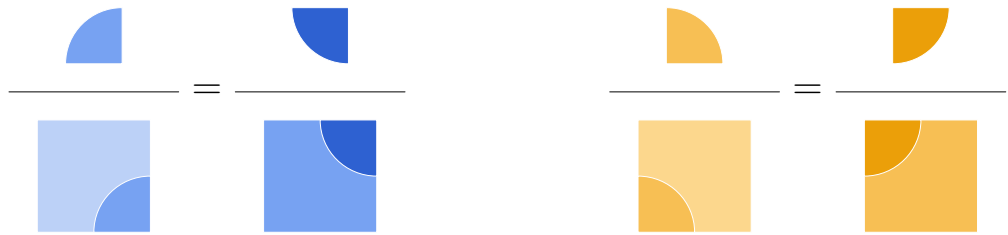

In unserem Beispiel würde das Modell entsprechend als fair bezeichnet werden, wenn
(a) die Wahrscheinlichkeit als tatsächlich für die Tätigkeit geeignete FLINTA* als geeignet klassifiziert zu werden genauso hoch ist wie die Wahrscheinlichkeit als tatsächlich geeigneter Mann als geeignet klassifiziert zu werden 
und 
(b) die Wahrscheinlichkeit als tatsächlich für die Tätigkeit ungeeignete FLINTA* als geeignet klassifiziert zu werden genauso hoch ist wie die Wahrscheinlichkeit als tatsächlich ungeeigneter Mann als geeignet klassifiziert zu werden. 
Eine Visualisierung dieser Gleichheiten ist in Abbildung \ref{figure_Gruppenfairness_EqualizedOdds} zu finden.

Lockerungen dieser Fairness-Definitionen sind \textit{Equal Opportunity} und \textit{Predictive Equality}, welche nur die Gleichheit einer der beiden Fehlerraten über die betrachteten Gruppen hinweg fordern: Equal Opportunity (vgl.\ Abb.\ \ref{figure_Gruppenfairness_EqualOpportunity}) fordert die Gleichheit der \glspl{RPR}, Predictive Equality (vgl.\ Abb.\ \ref{figure_Gruppenfairness_PredictiveEquality}) die der \glspl{FPR} (vgl.\ \cite{Ruf2021RightKindOfFairness}).
Dies ist immer dann sinnvoll, wenn nur einer der beiden Fehlerraten von Relevanz ist (vgl. Abschnitt \ref{subsection_AuswahlVonFairness-Definitionen}). Allerdings ist zu beachten, dass das Ausgleichen des einen Fehlers über die Gruppen hinweg zu einer größeren Ungleichheit des verbleibenden Fehlers über die Gruppen hinweg führen kann (vgl.\ \cite{Pessach2022ReviewOnFairness}).

Einfach gesagt fordert Separation in unserem Erklärbeispiel die gleiche Chance bzw. Wahrscheinlichkeit für Männer und FLINTA* korrekt geeignet und/oder korrekt ungeeignet klassifiziert zu werden. Dies fordert in der Regel eine gute Modellierung der beobachteten Daten über alle betrachteten Gruppen hinweg.
\textcolor{white}{\\}

\noindent 
Warum es manchmal trotz des Lösens des Problems des Unabhängigkeitskriteriums durch Hinzuziehen der Label Sinn ergibt, eine der Fairness-Definitionen aus dem Unabhängigkeitskriterium zu verwenden, diskutieren wir in Abschnitt \ref{subsection_AuswahlVonFairness-Definitionen}. Nun aber widmen wir uns der letzten Unterkategorie der Gruppenfairness.

\subsubsection{Suffizienz}
\label{subsubsection_Suffizienz}

\begin{wrapfigure}{r}{0.3\textwidth}
    \vspace{-10pt}
    \centering
    \resizebox{0.55\linewidth}{!}{
    \begin{tikzpicture}[line cap=round,line join=round,>=triangle 45,x=1cm,y=1cm,fill opacity=1]

%grid
%\draw [color=grid,, xstep=0.5cm,ystep=0.5cm] (-2,-2) grid (2,2);

\begin{scope}[shift={(0,1.5)}]

%quarter circles (start angle : stop angle : radius)

% upper left
\filldraw[fill=TP_discriminated, draw=white] (0,0) -- (-1,0) arc (180:90:1) -- (0,0);

% lower left
\filldraw[fill=TP_privileged, draw=white] (0,0) -- (-1,0) arc (180:270:1) -- (0,0);
\end{scope}

\draw[color=black] (-2,0) -- (2,0);

\begin{scope}[shift={(0,-1.5)}]

%quarter circles (start angle : stop angle : radius)
 % upper right
\filldraw[fill=FP_discriminated, draw=white] (0,0) -- (1,0) arc (0:90:1) -- (0,0);
% upper left
\filldraw[fill=TP_discriminated, draw=white] (0,0) -- (-1,0) arc (180:90:1) -- (0,0);
% lower right
\filldraw[fill=FP_privileged, draw=white] (0,0) -- (1,0) arc (0:-90:1) -- (0,0);
% lower left
\filldraw[fill=TP_privileged, draw=white] (0,0) -- (-1,0) arc (180:270:1) -- (0,0);
\end{scope}

\end{tikzpicture}}
    \caption{Darstellung des \acrshort{PPW}, bereits inklusive Visualisierung der betrachteten Gruppen.}
    \label{figure_PPV}
    \vspace{-10pt}
\end{wrapfigure}
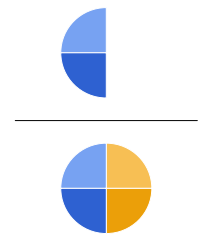
Das Konzept der Suffizienz unterscheidet sich zu dem der Separation anhand der zugrundeliegenden Grundgesamtheit: Während beim Konzept der Separation basierend auf der Menge aller Datenpunkte einer Klasse Statistiken über die Vorhersage des Modells berechnet werden, werden beim Konzept der Suffizienz basierend auf der Menge aller Datenpunkte, die eine gleiche Art der Vorhersage des Modells erhalten haben, Statistiken über die tatsächliche Klassenzugehörigkeit berechnet.
Typische, erneut von der Fairness-Fragestellung zunächst losgelöste Evaluationsmaße bzw.\ \textit{Fehlerraten} eines binären Klassifikators sind hier der \gls{PPW} sowie der \gls{NPW}.

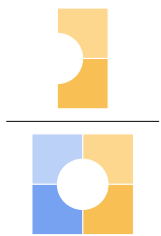
\begin{wrapfigure}{l}{0.3\textwidth}
    \vspace{-10pt}
    \centering
    \resizebox{0.45\linewidth}{!}{
    \begin{tikzpicture}[line cap=round,line join=round,>=triangle 45,x=1cm,y=1cm,fill opacity=1]

%grid
%\draw [color=grid,, xstep=0.5cm,ystep=0.5cm] (-2,-2) grid (2,2);

\begin{scope}[shift={(-1,2.5)} ]
%rectangles
%upper right
\filldraw[fill=TN_discriminated, draw=white] (0,0) rectangle (2,2);
%upper left
%\filldraw[fill=FN_discriminated, draw=white] (0,0) rectangle (-2,2); 
%lower right
\filldraw[fill=TN_privileged, draw=white] (0,0) rectangle (2,-2);
%lower left
%\filldraw[fill=FN_privileged, draw=white] (0,0) rectangle (-2,-2);

\begin{scope}[fill opacity=1]
%quarter circles (start angle : stop angle : radius)
 % upper right
\filldraw[fill=white, draw=white] (0,0) -- (0,1) arc (90:-90:1) -- (0,0);
% upper left
%\filldraw[fill=white, draw=white] (0,0) -- (-1,0) arc (180:90:1) -- (0,0);
% lower right
%\filldraw[fill=white, draw=white] (0,0) -- (1,0) arc (0:-90:1) -- (0,0);
% lower left
%\filldraw[fill=white, draw=white] (0,0) -- (-1,0) arc (180:270:1) -- (0,0);
\end{scope}
\end{scope}

\draw[color=black] (-3,0) -- (3,0);

\begin{scope}[shift={(0,-2.5)}]
%rectangles
%upper right
\filldraw[fill=TN_discriminated, draw=white] (0,0) rectangle (2,2);
%upper left
\filldraw[fill=FN_discriminated, draw=white] (0,0) rectangle (-2,2); 
%lower right
\filldraw[fill=TN_privileged, draw=white] (0,0) rectangle (2,-2);
%lower left
\filldraw[fill=FN_privileged, draw=white] (0,0) rectangle (-2,-2);

\begin{scope}[fill opacity=1]
%quarter circles (start angle : stop angle : radius)
 % upper right
\filldraw[fill=white, draw=white] (1,0) arc (0:360:1);
% upper left
%\filldraw[fill=white, draw=white] (0,0) -- (-1,0) arc (180:90:1) -- (0,0);
% lower right
%\filldraw[fill=white, draw=white] (0,0) -- (1,0) arc (0:-90:1) -- (0,0);
% lower left
%\filldraw[fill=white, draw=white] (0,0) -- (-1,0) arc (180:270:1) -- (0,0);
\end{scope}
\end{scope}
  
\end{tikzpicture}}
    \caption{Darstellung des \acrshort{NPW}, bereits inklusive Visualisierung der betrachteten Gruppen.}
    \label{figure_NPV}
    \vspace{-10pt}
\end{wrapfigure}
Der \gls{PPW} entspricht der Wahrscheinlichkeit, dass ein Datenpunkt, der in die positive Klasse klassifiziert wird, tatsächlich der positiven Klasse angehört (vgl.\ Abb.\ \ref{figure_PPV}). 
Die zugrundeliegende Grundgesamtheit entspricht hier dem Kreis, also den positiven Vorhersagen, aus unserem Schema aus Abbildung \ref{figure_MLGrundbegriffe}.
Der \gls{NPW} entspricht dagegen der Wahrscheinlichkeit, dass ein Datenpunkt, der in die negative Klasse klassifiziert wird, tatsächlich der negativen Klasse angehört (vgl.\ Abb.\ \ref{figure_NPV}).
In diesem Fall entspricht die zugrundeliegende Grundgesamtheit dem Bereich außerhalb des Kreises, also den negativen Vorhersagen, aus unserem Schema aus Abbildung \ref{figure_MLGrundbegriffe}.
Erneut führt das Unterteilen dieser zugrundeliegende Grundgesamtheiten in die betrachteten Gruppen gemäß dem Schema in Abbildung \ref{figure_MLGrundbegriffeInklusiveSensiblesMerkmal} (oben bzw.\ unten) zu verschiedenen Fairness-Definitionen:

Analog zu Equalized Odds im Separationskriterium fordert \textit{Conditional Use Accuracy Equality} gerade die Gleichheit der \glspl{PPW} sowie der \glspl{NPW} über die betrachteten Gruppen hinweg (vgl.\ \cite{Ruf2021RightKindOfFairness}).

\begin{figure}[htb]
    \begin{subfigure}[t]{.49\textwidth}
        \centering
        \resizebox{0.75\linewidth}
        {!}{\begin{tikzpicture}[line cap=round,line join=round,>=triangle 45,x=1cm,y=1cm,fill opacity=1]

%grid
%\draw [color=grid,, xstep=0.5cm,ystep=0.5cm] (-3,-6) grid (3,6);
\begin{scope}[shift={(-2,0)}]
\begin{scope}[shift={(0.5,0.5)}]
%quarter circles (start angle : stop angle : radius)
% upper left
\filldraw[fill=TP_discriminated, draw=white] (0,0) -- (-1,0) arc (180:90:1) -- (0,0);

\end{scope}

\draw[color=black] (-1.5,0) -- (1.5,0);

\begin{scope}[shift={(0,-1.5)}]

%quarter circles (start angle : stop angle : radius)
 % upper right
\filldraw[fill=FP_discriminated, draw=white] (0,0) -- (1,0) arc (0:90:1) -- (0,0);
% upper left
\filldraw[fill=TP_discriminated, draw=white] (0,0) -- (-1,0) arc (180:90:1) -- (0,0);

\end{scope}
\end{scope}

\node at (0,0) {\huge $=$};

\begin{scope}[shift={(2,0)}]
\begin{scope}[shift={(0.5,1.5)}]
%quarter circles (start angle : stop angle : radius)
% lower left
\filldraw[fill=TP_privileged, draw=white] (0,0) -- (-1,0) arc (180:270:1) -- (0,0);

\end{scope}

\draw[color=black] (-1.5,0) -- (1.5,0);

\begin{scope}[shift={(0,-0.5)}]

%quarter circles (start angle : stop angle : radius)
% lower right
\filldraw[fill=FP_privileged, draw=white] (0,0) -- (1,0) arc (0:-90:1) -- (0,0);
% lower left
\filldraw[fill=TP_privileged, draw=white] (0,0) -- (-1,0) arc (180:270:1) -- (0,0);

\end{scope}
\end{scope} 
  
\end{tikzpicture}}
        \caption{Visualisierung der Gruppenfairness-Definition Predictive Parity.}
        \label{figure_Gruppenfairness_PredictiveParity}
    \end{subfigure}
    \hfill
    \begin{subfigure}[t]{.49\textwidth}
        \centering
        \resizebox{0.75\linewidth}
        {!}{\begin{tikzpicture}[line cap=round,line join=round,>=triangle 45,x=1cm,y=1cm,fill opacity=1]

%grid
%\draw [color=grid,, xstep=0.5cm,ystep=0.5cm] (-3,-6) grid (3,6);
\begin{scope}[shift={(-3,0)}]
\begin{scope}[shift={(-1,0.5)}]
%rectangles
%upper right
\filldraw[fill=TN_discriminated, draw=white] (0,0) rectangle (2,2);
%upper left
%\filldraw[fill=FN_discriminated, draw=white] (0,0) rectangle (-2,2); 
%lower right
%\filldraw[fill=TN_privileged, draw=white] (0,0) rectangle (2,-2);
%lower left
%\filldraw[fill=FN_privileged, draw=white] (0,0) rectangle (-2,-2);

\begin{scope}[fill opacity=1]
%quarter circles (start angle : stop angle : radius)
 % upper right
\filldraw[fill=white, draw=white] (0,0) -- (0,1) arc (90:0:1) -- (0,0);
\end{scope}
\end{scope}

\draw[color=black] (-2.5,0) -- (2.5,0);

\begin{scope}[shift={(0,-2.5)}]

%rectangles
%upper right
\filldraw[fill=TN_discriminated, draw=white] (0,0) rectangle (2,2);
%upper left
\filldraw[fill=FN_discriminated, draw=white] (0,0) rectangle (-2,2); 
%lower right
%\filldraw[fill=TN_privileged, draw=white] (0,0) rectangle (2,-2);
%lower left
%\filldraw[fill=FN_privileged, draw=white] (0,0) rectangle (-2,-2);

\begin{scope}[fill opacity=1]
%quarter circles (start angle : stop angle : radius)
 % upper right
\filldraw[fill=white, draw=white] (0,0) -- (1,0) arc (0:180:1) -- (0,0);
% upper left
%\filldraw[fill=white, draw=white] (0,0) -- (-1,0) arc (180:90:1) -- (0,0);
% lower right
%\filldraw[fill=white, draw=white] (0,0) -- (1,0) arc (0:-90:1) -- (0,0);
% lower left
%\filldraw[fill=white, draw=white] (0,0) -- (-1,0) arc (180:270:1) -- (0,0);
\end{scope}
\end{scope}
\end{scope}

\node at (0,0) {\Huge $=$};

\begin{scope}[shift={(3,0)}]
\begin{scope}[shift={(-1,2.5)}]
%rectangles
%upper right
%\filldraw[fill=TN_discriminated, draw=white] (0,0) rectangle (2,2);
%upper left
%\filldraw[fill=FN_discriminated, draw=white] (0,0) rectangle (-2,2); 
%lower right
\filldraw[fill=TN_privileged, draw=white] (0,0) rectangle (2,-2);
%lower left
%\filldraw[fill=FN_privileged, draw=white] (0,0) rectangle (-2,-2);

\begin{scope}[fill opacity=1]
%quarter circles (start angle : stop angle : radius)
% lower right
\filldraw[fill=white, draw=white] (0,0) -- (1,0) arc (0:-90:1) -- (0,0);
\end{scope}
\end{scope}

\draw[color=black] (-2.5,0) -- (2.5,0);

\begin{scope}[shift={(0,-0.5)}]

%rectangles
%upper right
%\filldraw[fill=TN_discriminated, draw=white] (0,0) rectangle (2,2);
%upper left
%\filldraw[fill=FN_discriminated, draw=white] (0,0) rectangle (-2,2); 
%lower right
\filldraw[fill=TN_privileged, draw=white] (0,0) rectangle (2,-2);
%lower left
\filldraw[fill=FN_privileged, draw=white] (0,0) rectangle (-2,-2);

\begin{scope}[fill opacity=1]
%quarter circles (start angle : stop angle : radius)
 % upper right
\filldraw[fill=white, draw=white] (0,0) -- (1,0) arc (0:-180:1) -- (0,0);
% upper left
%\filldraw[fill=white, draw=white] (0,0) -- (-1,0) arc (180:90:1) -- (0,0);
% lower right
%\filldraw[fill=white, draw=white] (0,0) -- (1,0) arc (0:-90:1) -- (0,0);
% lower left
%\filldraw[fill=white, draw=white] (0,0) -- (-1,0) arc (180:270:1) -- (0,0);
\end{scope}
\end{scope}
\end{scope} 
  
\end{tikzpicture}}
        \caption{Visualisierung des zweiten Kriteriums für die Gruppenfairness-Definition Conditional Use Accuracy Equality.}
         \label{figure_Gruppenfairness_ConditionalUseCaseAccuracy_Condition2}
    \end{subfigure}
    \caption{Visualisierung der Gruppenfairness-Definition Conditional Use Accuracy Equality.}
    \label{figure_Gruppenfairness_ConditionalUseCaseAccuracy}
\end{figure}
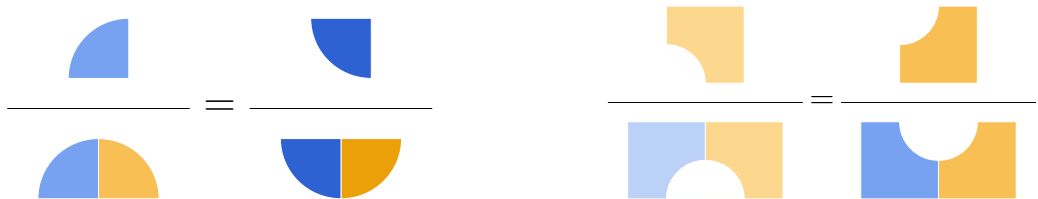

In unserem Beispiel würde das Modell entsprechend als fair bezeichnet werden, wenn 
(a) die Wahrscheinlichkeit als geeignet klassifizierte FLINTA* tatsächlich für die Tätigkeit geeignet zu sein genauso hoch ist wie die Wahrscheinlichkeit als geeignet klassifizierter Mann tatsächlich für die Tätigkeit geeignet zu sein
und
(b) die Wahrscheinlichkeit als nicht geeignet klassifizierte FLINTA* tatsächlich nicht für die Tätigkeit geeignet zu sein genauso hoch ist wie die Wahrscheinlichkeit als nicht geeignet klassifizierter Mann tatsächlich nicht für die Tätigkeit geeignet zu sein.
Eine Visualisierung dieser Gleichheiten ist in Abbildung \ref{figure_Gruppenfairness_ConditionalUseCaseAccuracy} zu finden.

Analog zu Equal Opportunity im Separationskriterium entspricht die Fairness-Definition \textit{Predictive Parity} einer Lockerung der Fairness-Definition Conditional Use Accuracy Equality, indem diese nur die Gleichheit der \glspl{PPW} über die betrachteten Gruppen hinweg fordert, wie in Abbildung \ref{figure_Gruppenfairness_PredictiveParity} dargestellt (vgl.\ \cite{Ruf2021RightKindOfFairness}). Während wir in der Literatur kein namentlich erwähntes Analogon im Suffizienzkriterium zur Predictive Equality im Separationskriterium gefunden haben, entspricht die Fairness-Definition in Abbildung \ref{figure_Gruppenfairness_ConditionalUseCaseAccuracy_Condition2} gerade eines solchen, indem es nur die Gleichheit der \glspl{NPW} fordert. Wir verzichten an dieser Stelle allerdings auf eine Namensgebung.

% Silbentrennung funktioniert nur so
Einfach gesagt fordert Suffizienz in unserem Erklärbeispiel die gleiche Wahrscheinlichkeit für Männer und FLINTA*, tatsächlich der Klasse zuzugehören, in welche sie klassifiziert wurden. Ein Nachteil des Suffizienzkriteriums kann sein, dass es durch Verschieben des Klassifika-tions-Schwellwerts manipulierbar ist: Wir können beispielsweise den \gls{PPW} für FLINTA* beliebig erhöhen, indem wir weniger FLINTA* anstellen, dafür aber nur die laut Modell am ehesten geeigneten. Das Separationskriterium hat diese Schwachstelle nicht.

\noindent 
Eine weitere Fairness-Definition, die etwas von der Grundidee abweicht, Fehlerraten wie die klassische \gls{RPR}, \gls{FPR} (vgl.\ Abschnitt \ref{subsubsection_Separation}), den \gls{PPW} oder den \gls{NPW} (vgl.\ dieser Abschnitt) über betrachtete Gruppen hinweg auszugleichen, ist der Begriff der \textit{Equal Calibration}, auch \textit{Calibration by Group} oder \textit{Test Fairness} genannt. Namensgebend ist hier das Konzept der Calibration, zu deutsch Kalibration, welches ebenfalls zunächst ein von der Fairness-Frage losgelöster Begriff der \gls{ML}-Theorie ist. 
Abweichend ist hier insbesondere die Annahme, dass nicht die finale Vorhersage des Klassifikators relevant ist, sondern der Wahrscheinlichkeitsscore\footnote{
    Während wir in Abschnitt \ref{subsection_Grundbegriffe} beliebige Scores zugelassen haben, ist der Begriff der Kalibration nur für Wahrscheinlichkeitsscores wohldefiniert.
} 
des Modells, also ein Wert zwischen 0 und 1 (vgl.\ Abschnitt \ref{subsection_Grundbegriffe}). 
Ein durch einen Wahrscheinlichkeitsscore ausgezeichnetes Modell ist dann kalibriert, wenn unter der Menge an Datenpunkten, die den Score $p \in [0,1]$ erhalten haben, die Wahrscheinlichkeit, tatsächlich zur positiven Klasse anzugehören, gerade wieder $p$ ist (vgl.\ \cite{Pleiss2017FairClassification_PostProcess_ModifyOutputs}). 
Unterteilt man die hier betrachtete Grundgesamtheit aller Datenpunkte mit selben Score erneut entsprechend der betrachteten Gruppen, erhält man die Fairness-Definition Equal Calibration. Diese ist also dann erfüllt, wenn gegeben eines vorhergesagten Wahrscheinlichkeitsscores $p \in [0,1]$, die Wahrscheinlichkeiten, dass eines unter diesen Exemplaren mit diesem Wahrscheinlichkeitsscore $p$ tatsächlich der positiven Klasse angehört, über die Gruppen hinweg gleich sind (vgl.\ \cite{Mehrabi2021SurveyOnFairness}). Gemäß der allgemeinen Definition der Kalibrierung wird in manchen Werken zusätzlich verlangt, dass diese gleichen Wahrscheinlichkeiten gerade wieder dem gegebenen Wahrscheinlichkeitsscore $p$ entsprechen sollen (vgl.\ \cite{Pessach2022ReviewOnFairness}).

In unserem Erklärbeispiel bedeutet dies, dass unter der Menge aller männlichen bzw.\ allen FLINTA*-Bewerber*innen, die laut Vorhersage des Modells zu 80$\%$ für den Beruf geeignet sind, tatsächlich 80$\%$ dieser männlichen bzw.\ FLINTA*-Bewerber*innen für den Beruf geeignet sind.
\\
\\
Abschließend ist noch zu erwähnen, dass in der Praxis die exakte Gleichheit zwischen statistischen Maßen wie die Erfolgsraten in Abschnitt \ref{subsubsection_Unabhängigkeit} oder die Fehlerraten in Abschnitt \ref{subsubsection_Separation} und diesem Abschnitt über die Gruppen hinweg seltenst exakt erreicht wird. Stattdessen versucht man in der Praxis, zumindest ähnliche Größen pro Gruppe zu erhalten. Dazu betrachtet man in der Regel die absolute Differenz zwischen den statistischen Maßen pro betrachteter Gruppe und versucht diese zu minimieren (vgl.\ \cite{Pessach2022ReviewOnFairness}).

\subsection{Individuelle Fairness}
\label{subsection_IndividuelleFairness}

Die zuvor genannten Definitionen der Gruppenfairness beruhen auf Statistiken, die über eine Menge an Datenpunkten hinweg betrachtet werden. Die betroffene Person interessiert allerdings vor allem, ob eine \acrshort{KI}-gestützte Entscheidung über sie im einzelnen fair ist. Zum Beispiel könnte sich eine FLINTA*-Bewerberin fragen, ob sie nicht trotz geltender Disparate Impact-Regelung abgelehnt wurde, weil sie zur betroffenen Gruppe gehört oder vielleicht auch gerade wegen der geltenden Regelung, da der festgelegte Anteil an FLINTA* bereits erreicht wurde.

So bieten die Gruppenstatistiken Spielraum, als unfair wahrgenommene Entscheidungen über Individuen zu treffen (vgl.\ \cite{Dwork2012}). Die Gruppenfairness-Definitionen treffen beispielsweise keine Aussage darüber, \textit{welche} Mitglieder der geschützten Gruppe ausgewählt werden sollen. Damit besteht Raum für böswillige Praktiken wie die selbst-erfüllende Prophezeiung.
In unserem Erklärbeispiel kann das beispielsweise bedeuten, dass zwar genügend FLINTA* als geeignet klassifiziert werden, aber spezifisch nur solche mit wenig Eignung, sodass sie nach dem Bewerbungsgespräch abgelehnt werden und in der Zukunft als warnendes Beispiel gegen Fairness-Regelungen herangezogen werden können.
Außerdem bieten die Gruppenfairness-Maße Raum für sogenanntes \textit{Subset Targeting}, bei der zwar nicht die gesamte geschützte Gruppe benachteiligt wird, aber ein bestimmter Teil davon.
Beispielsweise könnten gezielt Personen aus der FLINTA*-Gruppe mit Kindern nicht als geeignet klassifiziert werden, solange Erfolgs- oder Fehlerraten für die Gesamtmenge der FLINTA* eingehalten werden.
Unter das Subset Targeting fällt auch Intersektionalität.
So ist es möglich, genügend Schwarze Männer und \textit{weiße} FLINTA*-Bewerber*innen als geeignet zu klassifizieren und trotzdem gegen Schwarze FLINTA*-Bewerber*innen zu diskriminieren. Jede betrachtete Gruppe und jede Überschneidung, die in der Gruppenfairness betrachtet werden soll, muss explizit definiert werden (vgl.\ \cite{Castelnovo2022ClarificationOnFairness}).

Der Fokus der Methoden der individuellen Fairness liegt darauf, bewusste Entscheidungen über den Aus- oder Einschluss von Merkmalen zu treffen, um den Einfluss der sensiblen Merkmale zu eliminieren oder zu verringern.
\\
\\
\noindent
\textit{\gls{FTU}} oder auch \textit{Blindness} setzt die naheliegende Idee um, Merkmale, die wir als irrelevant für die Entscheidung erachten, einfach aus den Daten zu streichen (vgl.\ \cite{Castelnovo2022ClarificationOnFairness,Dwork2012}).
Die Annahme ist, dass ein Klassifikator, der nicht weiß, dass es sich um eine FLINTA*-Bewerber*in handelt, auch nicht diskriminieren kann. 
Allerdings bietet \gls{FTU} keinen Schutz gegen indirekte Diskriminierung.
Zudem kann auch direkte Diskriminierung weiterhin auftreten, da das Modell die Gruppenzugehörigkeit aufgrund anderer Merkmale, die das sensible Merkmal redundant enthalten, ermitteln kann (vgl.\ \cite{Dwork2012}), wie beispielsweise in dem in Abschnitt \ref{subsection_BeispieleFürUnfairness} genannten Beispiel des Bewerber*innen-Screenings von Amazon. Obwohl das Geschlecht nicht explizit angegeben war, haben Lebensläufe mit Sprache, die eher weiblich konnotiert war, zu einem eher negativen Score geführt (vgl.\ \cite{Dastin2018}). 

Weiterführende Ansätze versuchen, die Gruppenzugehörigkeit aus allen nicht sensiblen Merkmalen zu entfernen, indem sie eine faire Repräsentation dieser berechnen, die unabhängig von den sensiblen Merkmalen sind. Hier besteht aber vor allem die Herausforderung, die Daten dabei nicht zu sehr zu verfälschen, damit der Klassifikator nicht an Nutzen verliert  (vgl.\ \cite{Castelnovo2022ClarificationOnFairness}).
\\
\\
\noindent
Die Idee zu \textit{\gls{FTA}} von \cite{Dwork2012} leitet sich direkt komplementär zu \gls{FTU} ab.
Anstatt irrelevante Merkmale auszuschließen, soll hier explizit eingeschlossen werden, was wichtig ist, um Individuen für einen bestimmten Anwendungsfall zu vergleichen. 
Der zugrundeliegende Gedanke lautet \enquote{ähnliche Individuen müssen ähnlich behandelt werden}.
Dabei ergeben sich die zwei Fragen, was eine \textit{ähnliche Behandlung} und was \textit{ähnliche Individuen} bedeuten.
Beides gilt es, für einen spezifischen Klassifikator zu definieren.

\begin{figure}[htb]
    \begin{subfigure}[t]{.48\textwidth}
        \centering
        \resizebox{!}{3.5cm}
        {\begin{tikzpicture}[line cap=round,line join=round,>=triangle 45, label distance=15mm]

\filldraw [line width=1pt, draw=black, fill=TP_discriminated,opacity=1] (0,0) circle (2.5);
\begin{scriptsize}
\draw [line width=1pt, <->, color = black] (-1.5,0)-- node [below]{d} (1,1.5);
\draw [fill=white] (-1.5,0) circle (2.5pt); 
\node at (-1.5,-0.2) {Individuum A};

\draw [fill=white] (1,1.5) circle (2.5pt);
\node at (1,1.7) {Individuum B};

\node at (0,-1.2)[align=center] {\footnotesize Merkmalsraum};

\end{scriptsize}
\end{tikzpicture}}
        \caption{Abstand $d$ zwischen zwei Individuen im Merkmalsraum. Die Funktion zur Berechnung von $d$ muss kontextspezifisch definiert werden. Dabei sollen nur relevante Merkmale genutzt, sinnvoll skaliert und bei Bedarf die sensiblen Merkmale für normative Maßnahmen zur Hand genommen werden.}
        \label{fig:SimilarIndividuals}
    \end{subfigure}
    \hfill
    \begin{subfigure}[t]{.48\textwidth}
        \centering
        \resizebox{!}{3.5cm}
        {\definecolor{barcolor}{rgb}{0.2,0.2,1}

\begin{tikzpicture}[line cap=round,line join=round,>=triangle 45]
\begin{axis}[
ymajorgrids=true,
ymin=0,
ymax=1,
xmin=0,
xmax=1,
bar width = 20pt,
ytick={0,0.2,...,1},
ylabel=Score,
ytick align=outside, 
ytick pos=left,
xtick={0.3,0.7},
xticklabels={Individuum A,Individuum B},
major x tick style = transparent,
]    
\addplot[ybar,fill = TP_discriminated,opacity=1] coordinates {
    (0.3,0.6) 
    (0.7,0.8) 
};
\draw [line width=0.5pt,<->] (0.5,0.6)-- (0.5,0.8);

\begin{scriptsize}
\draw[color=black] (0.55,0.7) node {D};
%\draw [line width=0.5pt,<->] (0.4,0.6) -. (0.4,0.75);
%\draw[color=black] (0.45,0.7) node {d};
\end{scriptsize}
\end{axis}
\end{tikzpicture}}
        \caption{Abstand $D$ zwischen der Behandlung zweier Individuen. Für die Berechnung von $D$ werden in der Regel Standard-Distanzmaße zwischen Wahrscheinlichkeiten oder skalaren Scores benutzt.}
        \label{fig:SimilarTreatment}
    \end{subfigure}
    
    \caption{Das Lipschitz-Kriterium für individuelle Fairness: Der Unterschied $D$ der Behandlung ist beschränkt durch den Unterschied $d$ zwischen den Individuen (vgl.\ \cite{Dwork2012}).}
    \label{figure_LipschitzKriterium}
\end{figure}
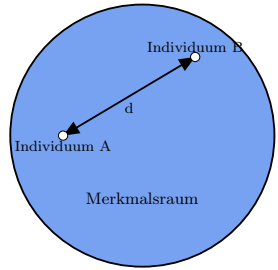
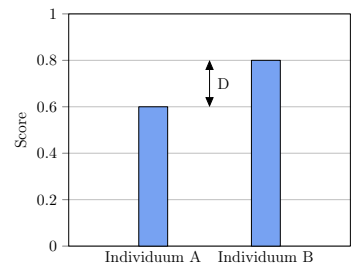

Für die ähnliche Behandlung verwendet man in der Regel nicht die finale binäre Vorhersage des Klassifikators, sondern den (Wahrscheinlichkeits-)Score des Modells (vgl.\ Abschnitt \ref{subsection_Grundbegriffe}). Mithilfe dessen können sowohl kleine als auch große Unterschiede in der Behandlung zweier Individuen festgestellt werden, nämlich durch den Unterschied derer prädizierten Scores (vgl.\ Abb.\ \ref{fig:SimilarTreatment}).

\label{sec:fta}

\noindent
Die Definition von Ähnlichkeit zwischen Individuen bietet hingegen deutlich mehr Spielraum und ist dadurch auch komplexer.
Hierfür ist es notwendig, den Abstand zwischen zwei beliebigen Individuen im Raum der Merkmale zu bemessen (vgl.\ Abb.~\ref{fig:SimilarIndividuals}). Allerdings können hierbei nicht einfach bekannte Distanzmaße auf dem gesamten Merkmalsraum angewandt werden, da verschiedene Merkmale unterschiedlich skaliert sind. Im Erklärbeispiel kann ein Notendurchschnitt beispielsweise nicht einfach auf die Anzahl der Jahre an Berufserfahrungen addiert werden. Außerdem können unterschiedliche Ausprägungen bei bestimmten Merkmalen unterschiedliche Wichtigkeit für die Ähnlichkeit von Individuen bedeuten. So können verschiedene Berufsabschlüsse eine ähnliche Qualifizierung ausdrücken oder eine Abschlussnote besonders wichtig oder unwichtig für verschiedene Arbeitgeber*innen sein.
Demnach muss ein Ähnlichkeitsmaß für jeden Kontext definiert werden.

Das gesamte individuelle Fairness-Kriterium ergibt sich aus der Kombination der beiden Ähnlichkeits- bzw.\ Abstandsbegriffe: Der Unterschied in der Behandlung jeder zwei Individuen darf nicht größer sein als ein Vielfaches des Unterschieds der Individuen selbst. Wie viel des Vielfachen der Ähnlichkeit der Individuen erlaubt ist obliegt dem Ersteller der \acrshort{KI}-Methode. Die resultierende Bedingung nennt sich \textit{Lipschitz-Kriterium} (vgl.\ Abb.\ \ref{figure_LipschitzKriterium}).

Durch den Spielraum für die Definition der Ähnlichkeit zwischen Individuen ergeben sich sowohl die Vorteile als auch die größten Nachteile dieses Ansatzes. Einerseits kann ein Ähnlichkeitsmaß formuliert werden, das Leistungen vergleicht ohne das Geschlecht einzubeziehen. Andererseits könnte man die \enquote{Awareness} des Ansatzes nutzen, um normative Ansätze zu verfolgen. 
So könnte in unserem Erklärbeispiel die Verwendung eines ausgleichenden Bonus, welcher zur Berufserfahrung der FLINTA* zugerechnet wird, um mögliche, durch längere Care-Arbeit entstehende Berufspausen auszugleichen, genutzt werden. Wenn solche Einzelfaktoren aber bekannt und diese auch beispielsweise in der Bewerbung eingetragen sind, kann auch gruppenunabhängig darauf eingegangen werden.
Ein solcher normativer Eingriff könnte dann hauptsächlich der FLINTA*-Gruppe zugutekommen, aber generell für alle Bewerber*innen gelten. Deshalb kann so auch Subset Targeting verhindert und intersektionale Fairness aufgebaut werden, da auch benachteiligte Bewerber*innen ohne Zugehörigkeit zu einer der geschützten Gruppen berücksichtigt werden können. Ein manuelles Maß zu formulieren, das Faktoren einzeln gewichtet und ggf.\ normativ korrigiert, erfordert jedoch viel Mühe und Expertise.

Vorgehensweisen, die versuchen das Ähnlichkeitsmaß automatisch zu finden, weisen wieder altbekannte Probleme auf. Zum Beispiel könnte man Expert*innen Beispiel-Paare bewerten lassen und daraus ein Distanzmaß erlernen. Dadurch läuft man allerdings Gefahr, menschlichen und historischen Bias ins System einzuführen.

Eine andere Möglichkeit ist es ein reines Leistungsmaß auf den nicht-geschützten Individuen zu definieren und andere Gruppen damit zu vergleichen, also beispielsweise den höchsten Score einer FLINTA*-Bewerber*in als gleichwertig zum höchsten Score eines männlichen Bewerbers zu betrachten. Diese Betrachtungsweise ähnelt sich wiederum der Gruppenfairness und die Vorteile von Intersektionalität oder der Gleichbehandlung von nicht-geschützten benachteiligten Individuen können wieder verloren gehen.

\subsection{Kausale Fairness}

Die bisher behandelten Methoden der Gruppen- sowie individuellen Fairness basieren auf beobachteten stichprobenbasierten Repräsentationen der Grundgesamtheit. Bei kausalen Modellen hingegen wird eine Repräsentation des Weltausschnitts basierend auf Expert*in\-nen\-wis\-sen zu Ursachen und Effekten erstellt. Dazu gehört zum einen ein kausaler Graph und zum anderen strukturelle Gleichungen. 

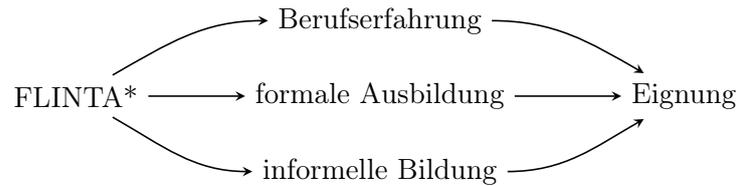
\begin{figure}[htb]
\centering
\begin{tikzpicture}
\node (weiblich)  at (-4,0) {FLINTA*};
\node (erfahrung) at (0,+1) {Berufserfahrung};
\node (formale_ausbildung) at (0, 0) {formale Ausbildung};
\node (informelle_bildung) at (0,-1) {informelle Bildung};
\node (eignung) at (+4,0) {Eignung};

\path[semithick, >=stealth, ->]
(weiblich) edge[out=30,in=180] (erfahrung)
(weiblich) edge[out=0,in=180] (formale_ausbildung)
(weiblich) edge[out=-30,in=180] (informelle_bildung)
(erfahrung) edge[out=0,in=150] (eignung)
(formale_ausbildung) edge[out=0,in=180] (eignung)
(informelle_bildung) edge[out=0,in=-150] (eignung);

\end{tikzpicture}
\caption{Illustration eines kausalen Modells für unser Erklärbeispiel aus Abschnitt~\ref{subsection_Grundbegriffe}. Unser kausales Modell nimmt an, dass Eignung von Berufserfahrung, formaler Ausbildung und informeller Bildung beeinflusst wird.
Alle drei Faktoren werden allerdings ihrerseits beeinflusst davon, ob Bewerber*innen FLINTA* sind oder nicht.}
\label{fig:causal_model}
\end{figure}

In dem Graph werden die Merkmale des Anwendungsfalls als Knoten dargestellt und auf Basis von Expert*innenwissen mit gerichteten Kanten verbunden, sodass die Pfeile die kausalen Beziehungen zwischen den Merkmalen widerspiegeln (vgl. \cite{Castelnovo2022ClarificationOnFairness,zhang2018causal}).
In Abbildung \ref{fig:causal_model} ist ein Graph für unser laufendes Beispiel abgebildet. Ob Bewerber*innen der FLINTA* Gruppe zugehörig sind hat Einfluss auf ihre Bildung, aber nicht andersherum.

\noindent
In nicht-kausalen, auf Beobachtungen basierenden Modellen kann hingegen nur Korrelation festgestellt werden, also dass FLINTA*-Sein und Bildung zusammenhängen, aber nicht in welche Richtung der Effekt auftritt.

Die strukturellen Gleichungen sind dazu da, um den genauen Effekt von Merkmalsausprägungen auf die danach folgenden Knoten zu berechnen.
Dabei werden auch "exogene" Knoten berücksichtigt, also solche Merkmale, die nicht erfasst werden (können). Dazu gehören in unserem Beispiel Kindheitserfahrungen der Bewerber*innen, die auf schwer nachvollziehbare Weise ihren Werdegang beeinflusst haben.
Diese exogenen Merkmale werden meist als Rauschen abgebildet, das auf die Gleichungen addiert wird (vgl. \cite{Castelnovo2022ClarificationOnFairness,zhang2018causal}).

Die Motivation für auf kausalen Modellen basierende Fairness ergibt sich aus sogenannten Counterfactuals, also hypothetische Fragen, die oft in Fairness-Szenarien auftreten. So kann man sich fragen, ob die Bewerber*in als geeignet klassifiziert worden wäre, wenn sie nicht-FLINTA* wäre. Hier wird jedoch nicht nur die Merkmalsausprägung von FLINTA* zu nicht-FLINTA* bzw.\ männlich getauscht, sondern auch die Effekte des Tausches auf die weiteren Merkmale berechnet. Auf Basis von kausalen Modellen wurden einige Fairness-Maße und -Methoden definiert, von denen wir uns eine Definition nun genauer ansehen (vgl. \cite{Castelnovo2022ClarificationOnFairness}).

Die Motivation der \textit{Path-specific Counterfactual Fairness} von \cite{zhang2018causal} ist es, Pfad-spezifisch auf Unfairness einzugehen.
Dabei ist ein Pfad ein Weg durch den Graphen von einem beliebigen Merkmal zum Ergebnis.
Ausgehend vom geschützten Merkmal wird für jeden Pfad ein Wert berechnet, der ausdrückt, wie viel der Pfad zum Label in den Datenpunkten oder der neuen Vorhersage des Modells beiträgt.
Die Berechnung erfolgt anhand von Counterfactuals zu Datenpunkten aus unserem observierten Datensatz, indem die Frage beantwortet wird, welchen Effekt es auf den Score hat, wenn die existierende Person eine männliche oder eine FLINTA* Bewerber*in wäre. In der Pfad-spezifischen Methode wird zusätzlich die Frage beantwortet, welchen Beitrag zur Änderung der spezifische Pfad geleistet hat.
Dabei sind Pfade vom geschützten Merkmal direkt zum Ergebnis direkte Diskriminierung, die es immer zu vermeiden gilt. Pfade, die vom geschützten Merkmal über andere Merkmale zum Ergebnis führen, können dagegen indirekte Diskriminierung ausdrücken. Allerdings kann die einzelne Betrachtung der Pfade dazu beitragen, Unfairness auf Pfaden zu erklären. So kann ein Pfad zwar als diskriminierend anerkannt, aber in diesem Kontext als notwendig betrachtet werden. 
Zum Beispiel könnte in unserem Graphen die gewünschte formale Ausbildung bei FLINTA* Bewerber*innen unwahrscheinlicher sein, aber informelle Bildung kann sie in diesem Fall nicht ersetzen, da es sich um eine zwingend notwendige Zertifizierung handelt. In dem Fall könnte entschieden werden, den diskriminierenden Effekt entlang des "formalen Ausbildung"-Pfades hinzunehmen.
Ein anderes Beispiel wäre die schlechtere Bezahlung von FLINTA*, die zum Teil durch reduzierte Arbeitsstunden begründet werden kann. Statt den Stundenlohn für FLINTA* zu erhöhen, kann der Pfad als gerechtfertigt betrachtet werden.
Für die Konstruktion der Counterfactuals führt eine solche Erklärung zu einem Durchtrennen des Pfads im Graphen. Für unser Erklärbeispiel heißt das: Wenn ein counterfactual Individuum die Ausprägung FLINTA* zugewiesen bekommt, wird dadurch die formale Ausbildung nicht mehr beeinflusst.

Der große Nachteil von kausalen Methoden besteht darin, das kausale Modell aufzustellen und zu verifizieren, denn
unterschiedliche kausale Modelle führen zu stark unterschiedlichen Schlussfolgerungen (vgl. \cite{Castelnovo2022ClarificationOnFairness}).

\subsection{Dynamische Fairness}
\label{subsection_DynamischeFairness}

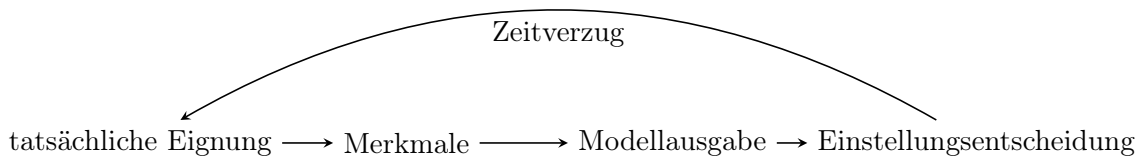
\begin{figure}[htb]
\centering
\begin{tikzpicture}
\node (y)  at (-5,0) {tatsächliche Eignung};
\node (x)  at (-1.5,0) {Merkmale};
\node (yhat) at (+2,0) {Modellausgabe};
\node (z) at (+6,0) {Einstellungsentscheidung};

\path[semithick, >=stealth, ->]
(y) edge (x)
(x) edge (yhat)
(yhat) edge (z)
(z) edge[out=150,in=30] node[below] {Zeitverzug} (y);

\end{tikzpicture}
\caption{Illustration eines dynamischen Modells für unser Erklärbeispiel aus Abschnitt~\ref{subsection_Grundbegriffe}.}
\label{fig:dynamic_model}
\end{figure}

Die bisher eingeführten Fairness-Definitionen betrachten Fairness im wesentlichen statisch: Ein Klassifikator ist zu einem gewissen Zeitpunkt fair oder nicht fair. Allerdings ist es auch denkbar, dass Klassifikatoren durch langfristige Effekte unerwünschte Nebeneffekte haben, die als unfair wahrgenommen werden. Betrachten wir unser laufendes Erklärbeispiel, so werden die laut Vorhersage des Modells geeignetsten Bewerber*innen eingestellt. Solange die geschätzte Eignung der tatsächlichen Eignung entspricht, würden die meisten der vorgenannten Definitionen dieses Schema als fair bezeichnen (vgl.\ \cite{CorbettDavies2018}).
Was passiert nun aber, wenn unsere Einstellungsentscheidung die Eignung \emph{in der Zukunft} beeinflusst (vgl.\ \cite{Perdomo2020})?
In unserem Beispiel könnte es etwa sein, dass FLINTA* in der Software-Entwicklung weniger vertreten sind und Software-Entwicklung deshalb für FLINTA* weniger attraktiv wird (z.B.\ aus Mangel an Vorbildern oder wegen einer sich herausbildenden, FLINTA*-feindlichen Fachkultur) und daher in den Folgejahrgängen sich noch etwas weniger FLINTA* dazu entscheiden, sich für den Job zu qualifizieren. Dementsprechend würde der Klassifikator noch weniger FLINTA* einstellen, was wiederum noch weniger qualifizierte FLINTA*-Bewerber*innen in der nächsten Generation bedeutet und so weiter. Es entsteht ein Teufelskreis, den man \emph{dynamisch unfair} nennt (vgl.\ \cite{Paassen2019}) und welcher in Abbildung \ref{fig:dynamic_model} visualisiert wird.

\noindent
Solche dynamischen Langzeiteffekte sind in der Praxis schwierig zu untersuchen, sodass die Forschung daran meist auf mathematische Modelle (vgl.\ \cite{creager2020causal,Liu2018}) und Simulationsstudien (vgl.\ \cite{Damour2020}) angewiesen ist. Unterschiedliche Modellannahmen ziehen dabei auch verschiedene Effekte nach sich (vgl.\ \cite{creager2020causal}). Gerade bei großflächigen Entscheidungsschemata (wie bei Einstellungspraktiken in Großunternehmen, Kredit-Scoring oder Hochschulzulassungen) sollten allerdings auch solche langfristigen Effekte bedacht werden, um ungewollte und Fairness-relevante Nebeneffekte auszuschließen.

\subsection{Auswahl von Fairness-Definitionen}
\label{subsection_AuswahlVonFairness-Definitionen}

Wie in den vorherigen Abschnitten deutlich wird, gibt es eine breite Auswahl an verschieden motivierten Fairness-Definitionen. Leider konnten in vorangegangenen Werken Konflikte zwischen diesen nachgewiesen werden, was eine Entscheidung für ein oder ein paar Fairness-Definitionen erfordert.
Beispielsweise beweisen \cite{Barocas2019FairnessBook}, dass die drei Konzepte Unabhängigkeit, Separation und Suffizienz der Gruppenfairness (vgl.\ Abschnitt \ref{subsection_Gruppenfairness}) in üblichen Fällen nicht gleichzeitig erfüllbar sind.\footnote{
    Eine Ausnahme stellt der Fall dar, dass Label und sensibles Merkmal tatsächlich unabhängig sind, was in der Regel auf Grund von historischem Bias nicht der Fall ist.}
\cite{Pleiss2017FairClassification_PostProcess_ModifyOutputs} weisen dagegen praktisch den Konflikt zwischen (verallgemeinerten) Gruppenfairness-Definitionen aus dem Bereich Separation und Suffizienz nach.
Ein intuitives Gedankenbeispiel liefern auch \cite{Pessach2022ReviewOnFairness} in Bezug auf das in Abschnitt \ref{subsection_BeispieleFürUnfairness} eingeführte COMPAS Modell:
Angenommen wir wollen voraussagen, ob ein Sträfling wieder straffällig wird, wobei dies statistisch betrachtet bei Männern eher der Fall ist als bei Frauen. 
Wenn wir Demographic Parity (Unabhängigkeit, vgl.\ Abschnitt \ref{subsubsection_Unabhängigkeit}) erreichen wollen, heißt das, dass das Modell mit gleicher Wahrscheinlichkeit die Rückfälligkeit von Männern wie von Frauen vorhersagen und damit mit gleicher Wahrscheinlichkeit Männern und Frauen Hafturlaub begünstigen soll.
Somit würde jedoch die Fehlerrate für fälschlicherweise prädizierte Rückfälligkeit für Frauen und die Fehlerrate für fälschlicherweise prädizierte nicht-Rückfälligkeit für Männer wachsen. Dann kann Equalized Odds (Separation, vgl.\ Abschnitt \ref{subsubsection_Separation}), also das Erreichen gleicher gewisser Fehlerraten für Männer und Frauen, nicht mehr erfüllt werden. 
Ist das Ziel stattdessen, Equalized Odds zu erfüllen, dann werden korrekterweise weniger Frauen als rückfällig vorhergesagt, als Männer, sodass Demographic Parity nicht erfüllt sein kann.

Ein wesentlicher Unterschied zwischen den hierbei genannten Konzepten ist das Verständnis von Fairness und das Vertrauen in den zugrundeliegenden Datensatz und vor allem, dessen Label (vgl.\ \cite{Barocas2019FairnessBook,Castelnovo2022ClarificationOnFairness,Ruf2021RightKindOfFairness,saleiro_aequitas_2019}). 
So besitzt das Konzept der Unabhängigkeit (vgl.\ Abschnitt \ref{subsubsection_Unabhängigkeit}) beispielsweise normativen Charakter: Nicht das Erreichen von (gleichermaßen) geringen Fehlerraten ist relevant, sondern das, was wir beispielsweise durch Richtlinien umsetzen wollen.
So können unter diesen Fairness-Definitionen höhere Fehlerraten zugunsten gleicher Erfolgsraten und zugunsten der betroffenen Gruppe erlaubt werden. Dies ist beispielsweise auch dann sinnvoll, wenn in der Realität die wahren Label zwar vom sensiblen Merkmal abhängen, der Ursprung dieser Ungleichheit allerdings Diskriminierung ist (vgl.\ \cite{Barocas2019FairnessBook,Paassen2019,Ruf2021RightKindOfFairness}). 
Die Konzepte der Separation sowie der Suffizienz (vgl.\ Abschnitt \ref{subsubsection_Separation} und \ref{subsubsection_Suffizienz}) ergeben dagegen immer dann Sinn, wenn die Label verfügbar und repräsentativ bzw.\ unverzerrt sind und die Annahme, dass Label und sensibles Merkmal in Relation stehen, sinnvoll ist (vgl.\ \cite{Castelnovo2022ClarificationOnFairness,Ruf2021RightKindOfFairness}). Ersteres ist beispielsweise in unserem Erklärbeispiel nicht der Fall: Hier sollen die Label ausdrücken, ob ein*e Bewerber*in tatsächlich für den Beruf geeignet ist. Allerdings ist es schwer zu bestimmen, ob eine Person wirklich für einen Beruf geeignet gewesen wäre, wenn die Person gar nicht erst die Chance dazu erhalten hat.
Welches der beiden Konzepte Separation und Suffizienz gewählt werden soll, hängt dann von der Wahl der Fehlerrate ab, was beispielsweise von der Folge der Vorhersage eines Modells abhängig ist (vgl.\ \cite{Castelnovo2022ClarificationOnFairness,Ruf2021RightKindOfFairness}). 
\cite{saleiro_aequitas_2019} unterscheiden hierbei beispielsweise zwischen strafenden und unterstützenden Folgen. So soll bei strafenden Folgen einer positiven Vorhersage Fokus auf Maße, die die \gls{FP} mit einbeziehen, gelegt werden, um zu verhindern, dass die falschen Individuen bestraft werden. Sind die positiven Vorhersagen hingegen unterstützend, empfehlen sie einen Fokus auf Maße, die die \gls{FN} enthalten, um möglichst wenigen Betroffenen die Hilfe zu verwehren, die sie benötigen. Die Autor*innen liefern dazu einen Entscheidungsbaum als Hilfe, sich zwischen den Maßen zu entscheiden.
Während \cite{Ruf2021RightKindOfFairness} in einem weiteren Entscheidungsbaum feinere Unterscheidung zwischen den einzelnen Gruppenfairness-Definitionen, vor allem zwischen Separation und Suffizienz, treffen, schließt der von \cite{saleiro_aequitas_2019} Gruppenfairness und kausale Fairness ein.

% Silbentrennung funktioniert nur so
Neben den Konflikten innerhalb der Gruppenfairness analysiert \cite{binns2020apparent} die (In-) Kompatibilität zwischen individueller Fairness und Gruppenfairness, wo er im Allgemeinen keinen Konflikt sieht.
Laut \cite{binns2020apparent} lassen sich alle Fairness-Definitionen auf einem Spektrum zwischen zwei Weltansichten ansiedeln: \enquote{We are all equal} (WAE) und \enquote{What you see is what you get} (WYSIWYG).
Bezogen auf unser Erklärbeispiel entspricht die Weltansicht WAE, dass Bewerber*innen aus den unterschiedlichen Gruppen sich nur unterscheiden, weil strukturelle Benachteiligung existiert. Im Idealfall hätten FLINTA* die gleichen Programmier-Vorerfahrungen oder Noten im technischen Bereich, wenn sie in der Kindheit ähnlich gefördert worden wären wie Männer. 
Deshalb gilt es hier den erfahrenen Nachteil, beispielsweise durch Nachteil-ausgleichende individuelle Fairness oder Gruppenfairness wie Demographic Parity auszugleichen. 
In der Weltansicht WYSIWYG würde hingegen angenommen werden, dass FLINTA* natürlicherweise die Unterschiede wie beispielsweise den Unterschied in der Programmiererfahrung in ihren Merkmalen ausprägen und deswegen auch nicht mehr FLINTA* eingestellt werden sollen. 
Für diese Weltanschauung eignet sich eine individuelle Fairness-Definition, das die Task-relevanten Merkmale wie Vorkenntnisse für alle Gruppen gleich berechnet oder eine Gruppenfairness-Definition wie Equalized Odds, bei denen die Fehlerraten gemessen werden.
Abschließend betonen \cite{Castelnovo2022ClarificationOnFairness} allerdings, dass das Einhalten einer Gruppenfairness-Definition die Verletzung der individuellen Fairness bedeuten \textit{kann} und umgekehrt.

\cite{Barocas2019FairnessBook} schließen zusammenfassend, dass die Inkompatibilitätsergebnisse kein \enquote{Artefakt der statistischen Entscheidungsfindung} (engl. \enquote{artifact of statistical decision making}) sind, sondern das \enquote{moralische Dilemma} (engl. \enquote{moral dilemmas}), welche Fairness-Definition die richtige ist, offenlegen.

Für einen von den reinen Fairness-Definitionen losgelösten Blick haben \cite{Tubella2023} das Framework ACROCPoLis entwickelt. Das Modellierungsframework kann als gemeinsame Sprache genutzt werden, um über Fairness zu sprechen. Es bildet sozio-technische und kontextuelle Aspekte ab, die für eine Fairness Frage relevant sind, wie beispielsweise die Stakeholder in einem Prozess, ihre Rollen sowie Relationen untereinander. Ähnlich wie die Unified Modeling Language (UML) könnte ACROCPoLis genutzt werden, um die Planung und das Design einer Fairness-relevanten Anwendung geordneter und transparenter zu gestalten. Die Motivation rührt auch daher, dass klassische, formelle und auf statistischen Maßen beruhende Fairness-Definitionen die Komplexität des sozialen und politischen Kontext außer Acht lassen, worauf wir tiefer in Abschnitt \ref{section_DiskussionUndFazit} eingehen.  Diese Komplexität soll mit Hilfe des Frameworks besser erfasst und dokumentiert werden. Es bietet also keine konkreten Empfehlungen für bestimmte Fairness-Maße, hilft aber, die Fairness für ein spezifisches Szenario zu evaluieren und somit eine geeignete Definition zu finden.

\section{Strategien zur Erreichung von Fairness}
\label{section_StrategienZurErreichungVonFairness}

Selbst wenn für einen konkreten Kontext eine Fairness-Definition ausgewählt ist, stellt sich die Frage, wie eine eventuell vorhandene Unfairness nun praktisch festgestellt sowie anschließend gemildert werden kann.
Dieser Frage widmen wir uns in diesem Abschnitt. Zum einen stellen wir sogenannte \textit{Audit Tools} vor, also Anwendungen, mit welcher Modelle oder die Ausgaben eines Modells auf Fairness geprüft werden können.
Zum anderen stellen wir etablierte Methoden vor, welche einen \gls{ML}-basierten Algorithmus fairer machen können. In das Training eines solchen Algorithmus kann an drei Stellen eingegriffen werden:
Zum einen kann das Modell vor dem Training durch die Bearbeitung des Trainingsdatensatzes fairer gemacht werden. Solche Methoden nennen sich \textit{Pre-Processing} Methoden.
\textit{In-Processing} Methoden sind dagegen Ansätze, in welchen das Modell durch die Modifizierung des Trainingsalgorithmus fairer gemacht wird.
In \textit{Post-Processing} Methoden wird dagegen ein bereits trainier\-tes Modell derartig modifiziert, dass es fairer wird (\cite{Barocas2019FairnessBook,Mehrabi2021SurveyOnFairness,Pessach2022ReviewOnFairness}). 
Wie bei der Wahl der Fairness-Definitionen gibt es auch hier keinen per se besten Ansatz. Die Wahl ist vielmehr von Einflussfaktoren wie beispielsweise die Verfügbarkeit der Label, der sensiblen Merkmale während der Evaluation des Modells sowie der Wahl der Fairness-Definition abhängig (vgl.\ \cite{Pessach2022ReviewOnFairness}).

\subsection{Audit Tools}

Ein erster Schritt, um faire \gls{ML}-Modelle umzusetzen, ist herauszufinden, für welche Gruppen und Fairness-Definitionen die betroffene Anwendung Probleme aufweist. Ebenso ist es wichtig, die Effekte nach Anwendung der Fairness-Methoden messen zu können. Um einen Überblick über verschiedene Fairness-Maße zu erhalten, eignen sich Audit-Tools.
Zwei Beispiele dafür sind Aequitas von \cite{saleiro_aequitas_2019} und AI Fairness 360 von \cite{bellamy_ai_2019}.
Aequitas richtet sich nicht nur an \acrshort{KI}-Entwickler*innen in Forschung und Industrie, sondern auch an potentielle externe Auditor*innen, die beauftragt werden könnten, um dem Klassifikator eine gewisse Fairness zu zertifizieren.
Dies liegt daran, dass kein Zugriff auf das Modell selbst benötigt wird, sondern nur ein Testdatensatz bestehend aus alleine den Labeln und Vorhersagen des Modells.
AI Fairness hingegen richtet sich eher an die Entwickler*innen als an externe Auditor*innen und bietet eine open-source Library, die nicht nur zur Auswertung der Fairness-Definitionen, sondern auch zum Anwenden von Fairnessmethoden auf Datensatz und Modell geeignet ist.

\subsection{Pre-Processing Methoden}

Falls in einem \gls{ML}-basierten System eine Form der Unfairness festgestellt wurde, beispielsweise mittels der zuvor vorgestellten Audit Tools, ist es eine Möglichkeit, das zugrundeliegende \gls{ML}-Modell auf einem modifizierten Trainingsdatensatz erneut zu trainieren. Die Modifizierung des Trainingsdatensatzes zielt dabei darauf ab, das darauf trainierte neue Modell fairer zu machen. 

\cite{Feldman2015FairClassification_PreProcess_ModifyFeatures} modifizieren die nicht-sensiblen Merkmale der Trainingsdaten beispielsweise, sodass es nicht mehr möglich ist, basierend auf den nicht-sensiblen Merkmalen Rückschlüsse auf die sensiblen Merkmale zu schließen. Dadurch wird verhindert, dass das Modell die sensiblen Merkmale durch die nicht-sensiblen lernen kann, sodass das Modell, was anschließend nur noch auf den nicht-sensiblen Daten trainiert wird, nicht mehr bezüglich der sensiblen Merkmale diskriminieren kann.
In unserem Erklärbeispiel könnten beispielsweise die Begriffe in der Bewerbung, die auf das nicht-männliche Geschlecht hinweisen, durch ein neutrales Wort ersetzt werden, sodass das Modell das Geschlecht nicht lernen kann.

Die folgenden zwei Alternativen beruhen auf der Annahme, dass Unfairness in das System gelangt, da es zu wenige Positivbeispiele in der geschützten Gruppe und zu wenige Negativbeispiele in der nicht-geschützten Gruppe innerhalb der Trainingsdaten gibt, also in unserem Erklärbeispiel zu wenige als geeignet gelabelte FLINTA* und zu wenige als ungeeignet gelabelte Männer, von denen das Modell lernen kann.

\cite{Kamiran2009FairClassification_PreProcess_ModifyLabels} ändern diesen Umstand, indem sie statt den nicht-sensiblen Merkmalen die Label der Trainingsdaten so ändern, dass positive Label innerhalb der geschützten Gruppe und negative Label innerhalb der nicht-geschützten Gruppe häufiger auftreten. Dabei werden nur solche Datenpunkte modifiziert, bzgl.\ dessen Entscheidung sich ein Modell ohnehin unsicher ist.\footnote{
    Unsicherheit bedeutet in diesem Fall, dass nur solche Stichproben betrachtet werden, die nahe an der sogenannten Entscheidungsgrenze eines Modells liegen.
}

In einer anderen Arbeit tun \cite{Kamiran2010FairClassification_PreProcess_PreferentialSampling} dies dagegen, indem sie die Trainingsdaten durch Resampling modifizieren:
Dazu duplizieren sie nicht nur
Datenpunkte aus der geschützten Gruppe mit positivem Label, also geeignete FLINTA*-Bewerber*innen, und 
Datenpunkte aus der nicht-geschützten Gruppe mit negativem Label, also ungeeignete männliche Bewerber,
sondern entfernen zusätzlich
Datenpunkte aus der geschützten Gruppe mit negativem Label, also ungeeignete FLINTA*-Bewerber*innen, und
Datenpunkte aus der nicht-geschützten Gruppe mit positivem Label, also geeignete männliche Bewerber, 
gemäß der obigen Überlegung. 
Erneut werden hier nur solche Datenpunkte dupliziert oder entfernt, bzgl.\ dessen Entscheidung sich ein Modell ohnehin unsicher ist.

Die betrachten Fairness-Definitionen sind hier ähnliche Formen von Disparate Impact (\cite{Feldman2015FairClassification_PreProcess_ModifyFeatures}) und Demographic Parity (\cite{Kamiran2009FairClassification_PreProcess_ModifyLabels} und \cite{Kamiran2010FairClassification_PreProcess_PreferentialSampling}), wobei jedoch die Vorhersage des Modells durch die Label ersetzt werden. Entsprechend ist eine Schwachstelle aller drei Methoden, dass sie Fairness nur für die Daten, nicht aber für das darauf trainierte Modell verifizieren.

\subsection{In-Processing Methoden}
\label{subsection_In-ProcessingMethoden}

Anstatt des Trainingsdatensatzes kann auch der Trainingsalgorithmus modifiziert werden. 
Beispiele dafür liefern \cite{Zafar2017FairClassification_InProcess_CovarianceConstraints}, \cite{Zafar2017FairClassification_InProcess_CovarianceConstraints_2} und \cite{Strotherm2023FairClassification_InProcess_CovarianceConstraints}, indem sie die Fehlerfunktion, welche typischerweise während des Trainings eines Modells optimiert wird, um eine Nebenbedingung ergänzen, welche die Gruppenfairness des Modells fördern soll. Genauer sorgt die Fehlerfunktion dafür, dass das Modell Vorhersagen trifft, die die Label möglichst gut abbilden. Die Nebenbedingung erlaubt aber nur solche Vorhersagen, die fair sind, und schränkt die möglichen Ausgaben des Modell damit ein. 

Die Fairness wird in allen Fällen durch ein Maß, der Kovarianz, approximiert, welches berechnet, wie sehr das sensible Merkmal, z.B.\ das Geschlecht, sowie die Vorhersage des Modells, z.B.\ die Eignungsvorhersage, einzelner Datenpunkte zusammenhängen. Bei Fairness, also der Unabhängigkeit dieser Größen, würde man keinen Zusammenhang erwarten, was sich in einer Kovarianz von Null widerspiegelt. 
Im Falle von \cite{Zafar2017FairClassification_InProcess_CovarianceConstraints} wird die Kovarianz basierend auf \textit{allen} Datenpunkten, also beispielsweise basierend auf allen Bewerber*innen, berechnet.

Im Falle von \cite{Zafar2017FairClassification_InProcess_CovarianceConstraints_2} werden zur Berechnung der Kovarianz dagegen nur \textit{missklassifizierte} Datenpunkte betrachtet; in unserem Beispiel fälschlicherweise als geeignet klassifizierte und/oder fälschlicherweise als ungeeignet klassifizierte Bewerber*innen. Dadurch wird nicht wie in \cite{Zafar2017FairClassification_InProcess_CovarianceConstraints} ein Stellvertreter für Disparate Impact (vgl.\ Abschnitt \ref{subsubsection_Unabhängigkeit}), sondern für Separation und Suffizienz (vgl.\ Abschnitt \ref{subsubsection_Separation} und \ref{subsubsection_Suffizienz}) - also Fehlerraten - definiert.

\cite{Strotherm2023FairClassification_InProcess_CovarianceConstraints} betrachten dagegen multiple sensible Merkmale, indem sie jeweils die Kovarianz der einzelnen sensiblen Merkmale sowie der Vorhersage des Modells betrachten. In unserem Erklärbeispiel könnte man dadurch nicht nur die Fairness zwischen FLINTA* und Männern erzwingen, sondern beispielsweise auch die zwischen Schwarzen und \textit{weißen} Bewerber*innen. Die genutzten Fairness-Definitionen sind Disparate Impact und Equal Opportunity, wobei beide Fairness-Definitionen für die in dieser Arbeit betrachteten Gruppen äquivalent sind.
Dies ist sonst nicht der Fall ist, da Disparate Impact zum Unabhängigkeitsprinzip (vgl.\ Abschnitt \ref{subsubsection_Unabhängigkeit}) und Equal Opportunity zum Separationsprinzip (vgl.\ Abschnitt \ref{subsubsection_Separation}) gehört (vgl.\ \cite{Barocas2019FairnessBook} und Abschnitt \ref{subsection_AuswahlVonFairness-Definitionen}).

Alternativ stellen \cite{Zafar2017FairClassification_InProcess_CovarianceConstraints} und \cite{Strotherm2023FairClassification_InProcess_CovarianceConstraints} auch Methoden vor, wie man die Gruppenfairness mittels eines Fairness-Stellvertreter oder einer verallgemeinerten Definition des Disparate Impacts für multiple sensible Merkmale direkt optimieren kann unter einer Nebenbedingung, die sicherstellt, dass die Performance des Klassifikators nicht zu stark von der maximal erreichbaren Performance abweicht.
Dies bedeutet, es wird direkt nach einem möglichst fairen Modell gesucht, welches männliche und FLINTA*-Bewerber*innen im Sinne des Disparate Impact gleich behandelt, während man aber nur solche Modelle erlaubt, die einer gewissen Korrektheit genügen.

Neben Methoden zur Optimierung von Gruppenfairness existieren auch Methoden, um individuelle Fairness sicherzustellen:
Die in Abschnitt \ref{sec:fta} eingeführte Definition für individuelle \gls{FTA} von \cite{Dwork2012} kann für faires Training eingesetzt werden, indem sie ebenfalls als Nebenbedingung bei der Optimierung eingesetzt wird. 
Unabhängig von der genutzten Fehlerfunktion wird also sichergestellt, dass ähnliche Bewerber*innen nach gegebenem Distanzmaß nicht zu unterschiedliche Scores bekommen.

Solche Fairness-Nebenbedingungen haben den Vorteil, dass sie von externen Organisationen vorgegeben und anhand eines Testdatensatzes geprüft werden können, ohne die Fehlerfunktion zu kennen, sodass diese bei Bedarf auch geheim gehalten werden kann. Zum Beispiel kann eine Firma auf historischen Daten auf das spezielle Ziel hintrainieren, Bewerber*innen einen hohen Score zu geben, die nicht nur geeignet sind, sondern auch eine möglichst Lange Firmenzugehörigkeit prognostiziert bekommen.
Das Vorgeben der Fairness-Nebenbedingung durch Externe ist besonders für die individuelle Fairness attraktiv, da so die Mühe geteilt werden kann, das Ähnlichkeitsmaß für einen Kontext zu definieren.

\subsection{Post-Processing Methoden}
\label{subsection_Post-ProcessingMethoden}

Als Alternative zur Modifizierung des Trainingsalgorithmus kann auch ein bereits trainiertes Modell zugunsten der Fairness modifiziert werden. 
Eine einfache Post-Processing-Methode für Demographic Parity ist es beispielsweise, ein trainiertes Scoring-Modell getrennt für betrachtete Gruppen anzuwenden und dann jeweils die höchsten Scores positiv zu klassifizieren, also beispielsweise die zehn laut Modell geeignetsten FLINTA*-Bewerber*innen und die zehn laut Modell geeignetsten männlichen Bewerber als geeignet zu klassifizieren. Diese Art des Post-Processing wird allerdings gerade im US-Kontext scharf kritisiert, weil es bedeutet, dass unterschiedliche Schwellwerte für unterschiedliche Gruppen verwendet werden und deshalb beispielsweise männliche Bewerber*innen als geeignet klassifiziert würden, wenn sie FLINTA* wären (vgl.\ \cite{CorbettDavies2018}).

\cite{Hardt2016FairClassification_PostProcess_RetrainModel} trainieren dagegen ein neues Modell basierend auf einem alten Modell, indem sie einen Fehler zwischen dem neuen und alten Modell optimieren und als Nebenbedingung in diesem Optimierungsproblem Equalized Odds einfordern. In unserem Beispiel wird ein unfaires Eignungsmodell also solange (und minimal) verändert, bis es bei männlichen und FLINTA*-Bewerber*innen gleiche (möglichst gute) \glspl{RPR} und \glspl{FPR} aufzeigt.

Dagegen modifizieren \cite{Pleiss2017FairClassification_PostProcess_ModifyOutputs} die Ausgaben des vortrainierten Modells, indem sie innerhalb der Gruppe, welche unter dem unfairen Modell bevorzugt wird, Informationen zurückhalten, stattdessen einige Vorhersagen randomisieren und somit die Performance innerhalb dieser Gruppe verschlechtern. 
Würde unser Beispielmodell also tendenziell männlichen Bewerbern einen höheren Score zuweisen, so würden manche Vorhersagen für Männer einfach randomisiert werden.
Die Fairness-Definition, welches die Autor*innen hierbei betrachten, ist eine gewichtete Summe der verallgemeinerten \gls{FPR} und \gls{FNR}, was im Grunde einer gewichtete Summe der Grundbausteine eines verallgemeinerten Equalized Odds entspricht, sowie Equal Calibration.

\subsection{Konflikt zwischen Fairness und Performance}

Zusammenfassend zielen die zuvor vorgestellten Methoden darauf ab, die Fairness eines \gls{ML}-System zu erhöhen. Wie zuvor beschrieben, schränken beispielsweise die dazu im Training genutzten Nebenbedingungen wie in Abschnitt \ref{subsection_In-ProcessingMethoden} und \ref{subsection_Post-ProcessingMethoden} oder die Ran\-do\-mi\-sier\-ung von Modell-Vorhersagen in Abschnitt \ref{subsection_Post-ProcessingMethoden} mögliche Ausgaben des Modells ein. Eine natürliche Folge dessen ist, dass die allgemeine Performance des Modells zugunsten der Fairness abnimmt. Diese Problematik wird Fairness-Accuracy Trade-Off, zu deutsch Fairness-Ge\-nau\-ig\-keits\--Kon\-flikt, genannt (vgl.\ \cite{Pessach2022ReviewOnFairness}).
Oft lässt sich dieser Trade-Off mittels eines wählbaren Parameters, im \gls{ML}-Jargon dann \textit{Hyperparameter} genannt, regulieren. Beispielsweise wird die in Abschnitt \ref{subsection_In-ProcessingMethoden} beschriebene Kovarianz durch einen Hyperparameter beschränkt. 
Je kleiner der Hyperparameter gewählt wird, so mehr Fairness wird gefordert, so mehr nimmt aber die Performance des Modells ab (vgl.\ \cite{Strotherm2023FairClassification_InProcess_CovarianceConstraints}).
In unserem Erklärbeispiel würde man also durch die Wahl des Hyperparameters den Grad der Fairness in Bezug auf Eignungvorhersagen in Relation zu dem Geschlecht bestimmen, mit sinkendem Hyperparameter aber allgemein häufiger auftretende falsche Vorhersagen des Modells in Bezug auf die vorgeschriebenen Label zulassen. Man würde also zulassen, dass Bewerber*innen häufiger als falsch qualifiziert bzw.\ unqualifiziert prädiziert werden, solange die Vorhersagen fair sind.

%\begin{figure}{r}{0.5\textwidth}
%    \vspace{-15pt}
%    \centering
%    \includegraphics[width=0.49\textwidth]{ACCandDICoherence_vertical.pdf}
%    \caption{Darstellung des Trade-Offs zwischen Modell-Performance, gemessen in Accuracy, und Modell-Fairness, hier gemessen in Disparate Impact, für verschiedene Fairness-steigernde Modelle. Die abgebildete Kurve entsteht durch verschiedene Wahlen von Hyperparametern während des fairen Lernens und wird auch Pareto-Front genannt.}
%    \label{figure_FairnessAccuracyTradeOff}
%    \vspace{-15pt}
%\end{figure}
%
%Leider kommen die Fairness steigernde Methode aus den vorherigen Abschnitten nicht ohne Nachteile. So führen das Einschränken der zu optimierenden Fehlerfunktion durch Nebenbedingungen während des Trainings oder die Randomisierung von Modell-Vorhersagen dazu, dass die Performance des Modells zugunsten der Fairness abnehmen kann. Diese Problematik wird Fairness-Accuracy Trade-Off, zu deutsch Fairness-Genauigkeits Konflikt genannt (vgl.\ \cite{Pessach2022ReviewOnFairness}). In vielen In-Process Methoden wie in denen aus Abschnitt \ref{subsection_InProcessing} können die hier eingeführten Hyperparameter genutzt werden, um diesen Fairness-Accuracy Trade-Off durch verschiedene Wahlen dieses Hyperparameters zu regulieren. In Abbildung ist eine Visualisierung dieses Trade-Offs aus den praktischen Ergebnissen aus \cite{Strotherm2023FairClassification_InProcess_CovarianceConstraints} ersichtlich.

\section{Anwendung auf den europäischen Kontext}
\label{section_AnwendungAufDenEuropäischenKontext}

Im weltweiten Vergleich ist der europäische Kontext durch besonderen Wert auf vertrauenswürdige \acrshort{KI} und entsprechende Regulierungen geprägt (vgl.\ \cite{BringasColmenarejo2022,Calvi2023,Kaminski2021,Wachter2021}).
Besonders hervorzuheben sind die \gls{DSGVO}, die Gleichbehandlungsnormen der europäischen Grundrechtecharta, der Digital Services Act und der entstehende AI Act. Darüber hinaus existieren nicht-bindende Richtlinien, die versuchen, eine 
spezifisch-europäische Perspektive auf vertrauenswürdige \acrshort{KI} zu formulieren, vor allem die \emph{Guidelines for Trustworthy AI} (vgl.\ \cite{EuropeanComission2019TrustworthyAI}).
In diesen Rechtsnormen und Richtlinien steht die Fairness allerdings eher im Hintergrund,
während besonders Transparenz und die informierte Entscheidung der Bürger*innen im Vordergrund steht.
So verlangt etwa der AI Act eine transparente Dokumentation der Trainingsdaten und die  \gls{DSGVO} eine
Erklärung automatischer Entscheidungen (vgl.\ \cite{BringasColmenarejo2022,Calvi2023,Kaminski2021,Wachter2021}).
Wir konzentrieren uns hier auf die Regeln, die sich mit Fairness befassen.

Zunächst legen die Gleichbehandlungsnormen der EU folgende sensible Merkmale fest: Geschlecht, ethnischer Hintergrund, Hautfarbe, genetische Merkmale, Religion, Behinderung, Alter, sexuelle Orientierung, politische Überzeugung, Sprache, Zugehörigkeit zu einer anerkannten nationalen Minderheit, Besitz, soziale Herkunft und Geburt (vgl.\ \cite{BringasColmenarejo2022}).
Die \gls{DSGVO} verlangt generell Gleichbehandlungen bei automatischen Entscheidungen, die Menschen betreffen, also insbesondere bei Systemen des \gls{ML} (vgl.\ \cite{Kaminski2021}). Es wird allerdings keine explizite Fairness-Definition benannt.

Die umfassendsten Regeln enthält der Entwurf des AI Acts. Dieses Gesetz soll allgemeine Regeln für die Entwicklung, Bereitstellung und den Betrieb von \acrshort{KI}-Systemen in der EU formulieren.
Kern des AI Acts ist die Unterscheidung in die Risiko-Klassen \enquote{begrenztes Risiko}, \enquote{generative KI}, \enquote{hohes Risiko} und \enquote{unannehmbares Risiko} (vgl.\ \cite{Calvi2023,Veale2021}).
Systeme in der letzten Klasse -- nämlich Systeme zur kognitiven Verhaltensmanipulation, biometrische Echtzeit-Identifikationssysteme und Systeme zum \enquote{social scoring} -- sollen grundsätzlich (also abgesehen von Ausnahmen) verboten werden, während Systeme in den anderen Klassen unter Regularien erlaubt werden sollen. Aus Fairness-Sicht ist besonders die Klasse der Hochrisiko-Systeme interessant, weil dort die meisten in der EU erlaubten \gls{ML}-Systeme eingeschlossen sind, die mit kritischen Entscheidungen zu tun haben, nämlich Systeme in den Anwendungsdomänen Bildung, Zugang zu öffentlichen Diensten und Leistungen, Zugang zu Beschäftigung und Selbständigkeit (also auch unser laufendes Erklärbeispiel), Strafverfolgung, Migration, Asyl und Grenzkontrollen sowie Rechtsprechung (vgl.\ \cite{EUParlament2023}).

Für solche Hochrisiko-Systeme schreibt der Entwurf des AI Act vor, dass die Trainingsdaten umfassend dokumentiert werden und Qualitätsansprüchen genügen müssen (vgl.\ auch das Konzept \enquote{data sheets for data sets} aus dem US-Kontext \cite{Gebru2021}).
Darüber hinaus muss das System auch im Betrieb permanent auf Bias überprüft werden,
wozu auch gehört, die Leistung des Systems für unterschiedliche demographische Gruppen zu dokumentieren -- ähnlich zu Gruppenfairness-Definitionen (vgl.\ \cite{BringasColmenarejo2022,Calvi2023,Veale2021}).
%Bemerkenswert ist, dass es \textcolor{red}{auf Grundlage der \gls{DSGVO}} für die meisten \acrshort{KI}-Systeme \emph{nicht} gestattet ist, die Zugehörigkeit zu geschützten Gruppen zu erheben.
%Allerdings stellt der AI Act den Bereisteller*innen von Hochrisiko-Systemen hierzu eine Ausnahmeregelung zwecks Bias- (bzw.\ Fairness-) Monitoring bereit (vgl.\ \cite{Veale2021}).

Insgesamt entsteht der Eindruck, dass sich europäische Institutionen nicht auf eine bestimmte Fairness-Definition und eine bestimmte Strategie zur Erreichung von Fairness festlegen, sondern sich bemühen, den spezifischen Anwendungskontext (z.B.\ Bildung, Arbeitsmarkt, etc.) und das entsprechende Risiko des Systems in den Fokus zu rücken.
Dieser Ansatz trägt den teilweise sehr unterschiedlichen Voraussetzungen und Anforderungen der jeweiligen Anwendungskontexte Rechnung -- allerdings bedeutet es auch Rechtsunsicherheit in der Umsetzung: Ein Unternehmen in der EU, das ein Hochrisiko-\acrshort{ML}-System einsetzen will, könnte kaum mit letzter Sicherheit sagen, welche Fairness-Definition und welche Fairness-Strategien hinreichend wären, um den jetzigen und zukünftigen Regeln zu entsprechen.
Vermutlich wird sich erst im praktischen Wechselspiel von Unternehmen, Gerichten, Politik, Journalismus und Zivilgesellschaft der nächsten Jahre herauskristallisieren, welche Fairness-Definitionen und -Strategien in welchem Anwendungskontext überzeugen.
Es ist daher Praktiker*innen (mindestens für Hochrisiko-Systeme) zu empfehlen, Fairness kontinuierlich und interdisziplinär zu betrachten, denn eine schlichte Berufung auf eine einzelne Fairness-Definition reicht nicht aus (vgl.\ \cite{Calvi2023}).

\section{Diskussion und Fazit}
\label{section_DiskussionUndFazit}

In diesem Kapitel haben wir den Stand der Forschung zu Fairness in \acrshort{ML}-Systemen zusammengefasst, von historischen Beispielen für Unfairness (Abschnitt~\ref{subsection_BeispieleFürUnfairness}) über die vielfältigen Definitionen von Fairness (Abschnitt~\ref{section_Fairness-Definitionen}), über Strategien zur Erreichung von Fairness in \acrshort{ML}-Systemen (Abschnitt~\ref{section_StrategienZurErreichungVonFairness}) bis hin zur Einbettung in den europäischen Kontext (Abschnitt~\ref{section_AnwendungAufDenEuropäischenKontext}).
Wir haben festgestellt, dass es trotz (oder gerade wegen) umfassender, inter- und transdisziplinärer Vorarbeiten längst keinen einheitlichen Begriff von Fairness gibt. Je nach eigenem Weltbild, Vertrauen in die Label und Anwendungskontext können unterschiedliche Fairness-Definitionen und Strategien zur Erreichung von Fairness überzeugend wirken.

Dementsprechend kann auch dieses Kapitel keinen umfassenden, allgemeingültigen Ratschlag geben, welche Fairness-Definition und -Strategie vorzuziehen ist. Stattdessen verweisen wir auf die Vorarbeiten, die dabei helfen sollen, die richtige Fairness-Definition und -Strategie \emph{für den jeweils konkreten Anwendungsfall} zu identifizieren (vgl.\ auch Abschnitt~\ref{subsection_AuswahlVonFairness-Definitionen}). \cite{Birhane2022} insbesondere mahnen an, Fairness nicht abstrakt sondern konkret zu diskutieren und explizit die Gruppe zu benennen, die geschützt werden soll. In anderen Worten: Der eigentliche Test für Fairness ist nicht die Erfüllung einer bestimmten Fairness-Definition, sondern ob die eigenen Maßnahmen tatsächlich und konkret der Gruppe helfen, der geholfen werden soll. Mit diesem Fokus rücken auch zusätzliche Fragen in den Fokus, die zu Fairness gehören: Etwa ob \emph{mit} oder \emph{über} die geschützte Gruppe entschieden wird, wer die (ökonomische, politische) Macht hat, zu bestimmen, dass ein \acrshort{ML}-System eingesetzt werden soll und so fort. Wir haben uns in dieser Arbeit bemüht, stets an einem konkreten Beispiel -- nämlich dem Eignungsscoring für Software-Entwicklung mit FLINTA* als geschützter Gruppe -- zu erklären, wie die jeweiligen Fairness-Definitionen und -Strategien funktionieren.

Zwar bildet dieses Kapitel den Stand der Forschung zu Fairness in Klassifikations- und Scoring-Systemen umfassend ab, allerdings gibt es auch Fairness-relevante Themen, die aus unserem Fokus herausfallen. \cite{Bolukbasi2016BiasedWordEmbeddings} etwa befassen sich mit der Fairness in Sprachmodellen. Solche Modelle neigen dazu, Stereotype aus den Trainingsdaten zu lernen, sodass zum Beispiel Worte für technische Berufe (\enquote{software engineer}, \enquote{machine learning researcher}, etc.) eher mit Worten wie \enquote{Mann} assoziiert werden als mit Worten wie \enquote{Frau} (vgl.\ \cite{Bolukbasi2016BiasedWordEmbeddings}). Die Fairness von Sprachmodellen ist im Zuge der Verbreitung von Sprachmodell-gestützten Werkzeugen wie ChatGPT zunehmend in den Fokus der Debatte gerückt und auch der europäische AI Act wurde im Bezug auf solche Art von Systemen nachgebessert (vgl.\ \cite{EUParlament2023}).

Abschließend lässt sich feststellen, dass eine rein technische und abstrakte Perspektive dem Thema Fairness nicht gerecht wird. Es braucht die interdisziplinäre Zusammenarbeit aus Technik- und Sozialwissenschaften, die Expertise aus der jeweiligen Anwendungsdomäne und die Perspektive der konkret betroffenen, geschützten Gruppe, um eine kontextadäquate Definition und Strategie für Fairness zu finden -- und dann über den gesamten Lebenszyklus eines Systems umzusetzen.

\section*{Danksagung}
Wir bedanken uns für die Finanzierung durch das \gls{ERC} im Rahmen des \gls{ERC} Synergy Grant \enquote{Water-Futures} (Finanzhilfevereinbarung Nr.\ 951424).

\bibliographystyle{apalike-german}
\bibliography{literature}
\end{document}